\documentclass{article} 
\usepackage{iclr2024_conference,times}

\usepackage{amsmath,amsfonts,bm}

\def\eqref#1{equation~\ref{#1}}

\def\1{\bm{1}}

\def\rvx{{\mathbf{x}}}
\def\rvy{{\mathbf{y}}}
\def\rvz{{\mathbf{z}}}

\def\rmC{{\mathbf{C}}}

\def\rmS{{\mathbf{S}}}

\DeclareMathAlphabet{\mathsfit}{\encodingdefault}{\sfdefault}{m}{sl}
\SetMathAlphabet{\mathsfit}{bold}{\encodingdefault}{\sfdefault}{bx}{n}

\newcommand{\E}{\mathbb{E}}

\newcommand{\Enc}[2]{{q_{\phi}(#1|#2)}}

\newcommand{\Dec}[2]{{p_{\theta}(#1|#2)}}
\newcommand{\KL}[2]{D_{\mathrm{KL}}\left[#1\|#2\right]}

\usepackage{booktabs, tabularx}       
\usepackage{graphicx}
\usepackage{multicol, multirow}
\usepackage{tikz}
\usetikzlibrary{bayesnet}
\usetikzlibrary{calc}
\usetikzlibrary{arrows.meta}
\usepackage{amssymb}
\usepackage{algorithm}
\usepackage{algpseudocode}
\usepackage{adjustbox}
\usepackage{array}
\usepackage{wrapfig}

\newenvironment{propfirst}[1]{\begin{proposition}\label{#1} 
}{\end{proposition}}
\newenvironment{propagain}[1]{\begin{trivlist}\label{#1_appendix}
\item[] \textbf{Proposition~\ref{#1}.}\em}{\end{trivlist}}
\usepackage{enumitem}
\usepackage{capt-of}
\usepackage{varwidth}
\newsavebox\tmpbox

\newtheorem{proposition}{Proposition}
\newcommand{\STAB}[1]{\begin{tabular}{@{}c@{}}#1\end{tabular}}

\usepackage{hyperref}
\usepackage{url}

\title{Discouraging posterior collapse in hierarchical Variational Autoencoders using context}

\author{Anna Kuzina \\
Vrije Universiteit Amsterdam\\
\And
Jakub M. Tomczak \\
Eindhoven University of Technology \\
}

\iclrfinalcopy 
\begin{document}

\maketitle

\begin{abstract}
Hierarchical Variational Autoencoders (VAEs) are among the most popular likelihood-based generative models. 
There is a consensus that the top-down hierarchical VAEs allow effective learning of deep latent structures and avoid problems like posterior collapse. 
Here, we show that this is not necessarily the case, and the problem of collapsing posteriors remains. 
To discourage this issue, we propose a deep hierarchical VAE with a context on top. 
Specifically, we use a Discrete Cosine Transform to obtain the last latent variable.
In a series of experiments, we observe that the proposed modification allows us to achieve better utilization of the latent space and does not harm the model's generative abilities. 
\end{abstract}

\section{Introduction}

Latent variable models (LVMs) parameterized with neural networks constitute a large group in deep generative modeling \citep{tomczak2022deep}. One class of LVMs, Variational Autoencoders (VAEs) \citep{kingma2013auto, rezende2014stochastic}, utilize amortized variational inference to efficiently learn distributions over various data modalities, e.g., images \citep{kingma2013auto}, audio \citep{van2017neural} or molecules \citep{gomez2018automatic}. One of the problems hindering the performance of VAEs is the \textit{posterior collapse}  \citep{wang2021posterior} when the variational posterior (partially) matches the prior distribution (e.g., the standard Gaussian distribution). The expressive power of VAEs could be improved by introducing a hierarchy of latent variables. The resulting hierarchical VAEs like ResNET VAEs \citep{kingma2016improved}, BIVA \citep{maaloe2019biva}, very deep VAE (VDVAE) \citep{child2020very} or NVAE \citep{vahdat2020nvae} achieve state-of-the-art performance on images in terms of the negative log-likelihood (NLL). Despite their successes, hierarchical VAEs could still suffer from the posterior collapse effect. As a result, the modeling capacity is lower, and some latent variables carry very little to no information about observed data.

In this paper, we take a closer look into the posterior collapse in the context of hierarchical VAEs. It was claimed that introducing a specific \textit{top-down} architecture of variational posteriors \citep{sonderby2016ladder, maaloe2019biva, child2020very, vahdat2020nvae} solves the problem and allows learning powerful VAEs. However, we can still notice at least partial posterior collapse, where some of the latent variables are completely ignored by the model. 
Here, we fill a few missing gaps in comprehending this behavior. We analyze the connection between posterior collapse and latent variable non-identifiability. 
By understanding the issue that lies in the optimization nature of the Kullback-Leibler terms, we propose to utilize a non-trainable, discrete, and deterministic transformation (e.g., Discrete Cosine Transform) to obtain informative top-level latent variables. Making the top latent variables highly dependent on data, we alter the optimization process. The resulting hierarchical VAE starts utilizing the latent variables differently. In the experiments, we show that our proposition achieves a different landscape of latent space. 

The contributions of the paper are the following:
\begin{itemize}[itemindent=0pt,leftmargin=15pt]
    \item We provide empirical evidence that the posterior collapse is present in top-down hierarchical VAEs (Section~\ref{subsec:empirical_collapse}).
    \item We extend the analysis of the posterior collapse phenomenon presented by \citep{wang2021posterior} to hierarchical VAEs (Section~\ref{subsec:post_collapse_and_non_ident}).
    \item We propose a way to discourage posterior collapse by introducing Discrete Cosine Transform (DCT) as a part of the variational posterior (Section~\ref{sec:proposed_approach}).
    \item In the experiments, we show that the proposed approach leads to better latent space utilization (Section \ref{sect:posterior_collapse}), more informative latent variables (Section \ref{sect:compression}) and does not harm the generative performance (Section \ref{sect:exp_image_generations}).
\end{itemize}

\section{Background}

\subsection{Variational Autoencoders} \label{sec:vae}

Consider random variables $\rvx \in \mathcal{X}^{D}$ (e.g., $\mathcal{X} = \mathbb{R}$). We observe $N$ $\rvx$'s sampled from the empirical distribution $q(\rvx)$. We assume that each $\rvx$ has $L$ corresponding latent variables $\rvz_{1:L} = (\rvz_1, \dots, \rvz_L), \rvz_l \in \mathbb{R}^{M_{l}}$, where $M_l$ is the dimensionality of each variable. We aim to find a latent variable generative model with unknown parameters $\theta$, $p_{\theta}(\rvx, \rvz_{1:L}) = \Dec{\rvx}{\rvz_{1:L}} p_{\theta}(\rvz_{1:L})$. In general, optimizing latent-variable models with non-linear stochastic dependencies is troublesome. A possible solution is an approximate inference in the form of variational inference \citep{jordan1999introduction} with a family of variational posteriors over the latent variables $\{\Enc{\rvz_{1:L}}{\rvx}\}_{\phi}$. This idea is exploited in Variational Auto-Encoders (VAEs) \citep{kingma2013auto, rezende2014stochastic}, in which variational posteriors are referred to as encoders. As a result, we optimize a tractable objective function, i.e., the Evidence Lower BOund (ELBO), over the parameters of the variational posterior, $\phi$, and a generative part, $\theta$:
\begin{equation}\label{eq:vae_elbo}
\E_{q(\rvx)} \left[\ln p_{\theta}(\rvx)\right] \geq \E_{q(\rvx)} \biggl[ \E_{\Enc{\rvz_{1:L}}{\rvx}}\ln \Dec{\rvx}{\rvz_{1:L}} - \KL{\Enc{\rvz_{1:L}}{\rvx}}{p_{\theta}(\rvz_{1:L})} \biggr] ,
\end{equation}
where $q(\rvx)$ is an empirical data distribution. Further, we use $q^{\text{test}}(\rvx)$ for the hold-out data.
\subsection{Top-down hierarchical VAEs} \label{sec:hierarchical_vae}

We propose to factorize the distribution over the latent variables in an autoregressive manner: $p_{\theta}(\rvz_1, \ldots, \rvz_L) = p_{\theta}(\rvz_L)\prod_{l=1}^{L-1}\Dec{\rvz_{l}}{\rvz_{l+1:L}}$, similarly to \citep{child2020very, maaloe2019biva, vahdat2020nvae}. Next, we follow the proposition of \citep{sonderby2016ladder} with the top-down inference model: $\Enc{\rvz_1, \ldots, \rvz_L}{\rvx} = \Enc{\rvz_L}{\rvx}\prod_{l=1}^{L-1}\Enc{\rvz_l}{\rvz_{l+1:L}, \rvx}$. This factorization was used previously by successful VAEs, among others, NVAE~\citep{vahdat2020nvae} and Very Deep VAE (VDVAE)~\citep{child2020very}. It was shown empirically that such a formulation allows for achieving state-of-the-art performance on several image datasets.


\section{An analysis of the \emph{posterior collapse} in hierarchical VAEs}\label{sec:posterior_collapse}

The \textit{posterior collapse} effect is a known problem of shallow VAEs when certain latent variables do not carry any information about the observed data. There are various methods to deal with this issue for VAEs, such as changing the parameterization \citep{dieng2019avoiding, he2019lagging}, changing the optimization or the objective \citep{alemi2018fixing, bowman2015generating, fu2019cyclical, havrylov2020preventing, razavi2019preventing}, or using hierarchical models \citep{child2020very, maaloe2017semi, maaloe2019biva, tomczak2018vae, vahdat2020nvae}. Here, we focus entirely on the hierarchical VAEs since the posterior collapse problem is not fully analyzed in their context. 

In practice, hierarchical VAEs usually require huge latent space with multiple latent layers to achieve good performance \citep{sonderby2016ladder, maaloe2019biva, child2020very, vahdat2020nvae}. However, as we show in our analysis, the actual number of used latent units in these models is relatively small. Therefore, it is still an open question about how to reduce the gap between the total size of the latent space and the actual number of latents used by these models. 

Following definition 1 in \citet{wang2021posterior}, we consider the posterior collapse as a situation where the true posterior is equal to the prior for a given set of parameters $\theta$. We can formulate this definition for a single stochastic layer of top-down hierarchical VAE as follows:
\begin{align}\label{eq:posterior_collapse_definition}
p_{\theta}(\rvz_l|\rvz_{l+1:L}, \rvx) = p_{\theta}(\rvz_l|\rvz_{l+1:L}).    
\end{align}
In practice, we deal with the variational posterior $q_{\phi}(\rvz_{l}|\rvz_{l+1:L},\rvx)$, which approximates the true posterior. Furthermore, it is common to identify the posterior collapse based on this approximate distribution \citep{burda2015importance, lucas2019understanding, sonderby2016ladder, van2017neural}. 
Both definitions are connected, yet not identical. 
We learn the posterior approximation by variational inference, and the ELBO (Eq.~\ref{eq:vae_elbo}) is maximized when the approximate posterior matches the true posterior, namely, 
$\KL{\Enc{\rvz_{1:L}}{\rvx}}{p_{\theta}(\rvz_{1:L}|\rvx)} =0$. 
Furthermore, the KL-divergence can be further decomposed into the following sum:
$
  D_{\mathrm{KL}}[q_{\phi}(\rvz_{1:L}|\rvx) || p_{\theta}(\rvz_{1:L}|\rvx)] 
    =\KL{\Enc{\rvz_L}{\rvx}}{p_{\theta}(\rvz_L|\rvx)} 
    + \sum_{l=1}^{L-1} \E_{q_{\phi}(\rvz_{l+1:L},\rvx)} \KL{\Enc{\rvz_l}{\rvz_{l+1:L}, \rvx}}{p_{\theta}(\rvz_l|\rvz_{l+1:L}, \rvx)}.$

Therefore, a collapsed true posterior distribution for the latent variable at the stochastic layer $l$ results in a collapsed variational posterior for this latent variable at the optimum. 
However, the collapse of the variational posterior distribution does not guarantee the collapse of the true posterior as it can be caused by a poor choice of the family of the variational distributions.
See Appendix \ref{appdx:variational_collapse} for an in-depth discussion. To this end, we assume that the family of variational posterior distribution is rich enough and use the variational posterior collapse as an indicator of true posterior collapse. Next, we discuss the metrics of the posterior collapse in more detail. 


\subsection{Measuring the posterior collapse}
We consider two metrics for assessing the posterior collapse in hierarchical VAEs. First, we compute the \textit{KL-divergence} for the $i$-th latent variable of the stochastic layer $l$:
\begin{align}\label{eq:kl_collapse_metric}
    \text{kl}_l^i &=  \E_{q^{\text{test}}(\rvx)} \E_{q_{\phi}(\rvz_{l+1:L}|\rvx)}\KL{\Enc{\rvz^i_l}{\rvz_{l+1:L}, \rvx}}{p_{\theta}(\rvz^i_l|\rvz_{l+1:L})}.
\end{align}
This quantity can be approximately computed using Monte Carlo sampling and gives us an estimate of the posterior collapse issue for each latent variable. 
Note that the KL-divergence term used in the ELBO \ref{eq:vae_elbo} equals the sum of these values over all latent variables $i$ and stochastic layers $l$. 

Second, we use \textit{active units}. This is a metric introduced in \citep{burda2015importance}, and it can be calculated for a given stochastic layer and a threshold $\delta$:
\begin{align}\label{eq:active_units}
    \text{A}_{l} &= \text{Var}_{q^{\text{test}}(\rvx)} \E_{q_{\phi}(\rvz_{l+1:L}|\rvx)} \E_{q_{\phi}(\rvz_l | \rvz_{l+1:L}, \rvx)}\left[\rvz_l\right], \\
    \text{AU} &= \frac{\sum_{l=1}^L \sum_{i=1}^{M_l}  \left[\text{A}_{l, i} > \delta \right] }{\sum_{l=1}^L M_l },
\end{align}
where $M_{l}$ is the dimensionality of the stochastic layer $l$, $\left[ P \right]$ is Iverson bracket, which equals to $1$ if $P$ is true and to $0$ otherwise.
Following \citep{burda2015importance}, we use the threshold $\delta = 0.01$. The higher the share of active units, the more efficient the model is in using its latent space. 

\subsection{Empirical Evidence of Posterior Collapse}\label{subsec:empirical_collapse}
 \begin{table}[t]
 \begin{varwidth}[b]{0.55\textwidth}
    \centering
    \caption{Posterior collapse metrics and NLL for the top-down hierarchical VAEs with various \textsc{latent space} sizes and with fixed model \textsc{size} (the total number of parameters).} 
    \vskip 10pt
    \label{tab:mnist_latent_size_au}
    \begin{tabular}{lllccc}
        \toprule
        \footnotesize{\multirow{2}{*}{\textsc{Size}}} & \footnotesize{\multirow{2}{*}{\textsc{L}}}
          & \footnotesize{\textsc{Latent}}& \footnotesize{\multirow{2}{*}{\textsc{AU}} } & \footnotesize{\multirow{2}{*}{\textsc{KL}}}    &\footnotesize{\multirow{2}{*}{\textsc{NLL}$\downarrow$}}\\
          & & \textsc{Space} & &  & \\
        \midrule
            676K & 4  & 490  & 38.3\% & 0.047 &79.6 \\
            624K & 6  & 735  & 37.9\% & 0.031 &78.8 \\
            657K & 8  & 980  & 33.5\% & 0.022 &78.3 \\
            651K & 10 & 1225 & 33.6\% & 0.018 &77.9 \\
        \bottomrule
    \end{tabular}
    \vskip -12pt
 \end{varwidth}%
 \hfill
\begin{minipage}[t]{0.4\textwidth}
\centering
        \includegraphics[width=0.9\linewidth]{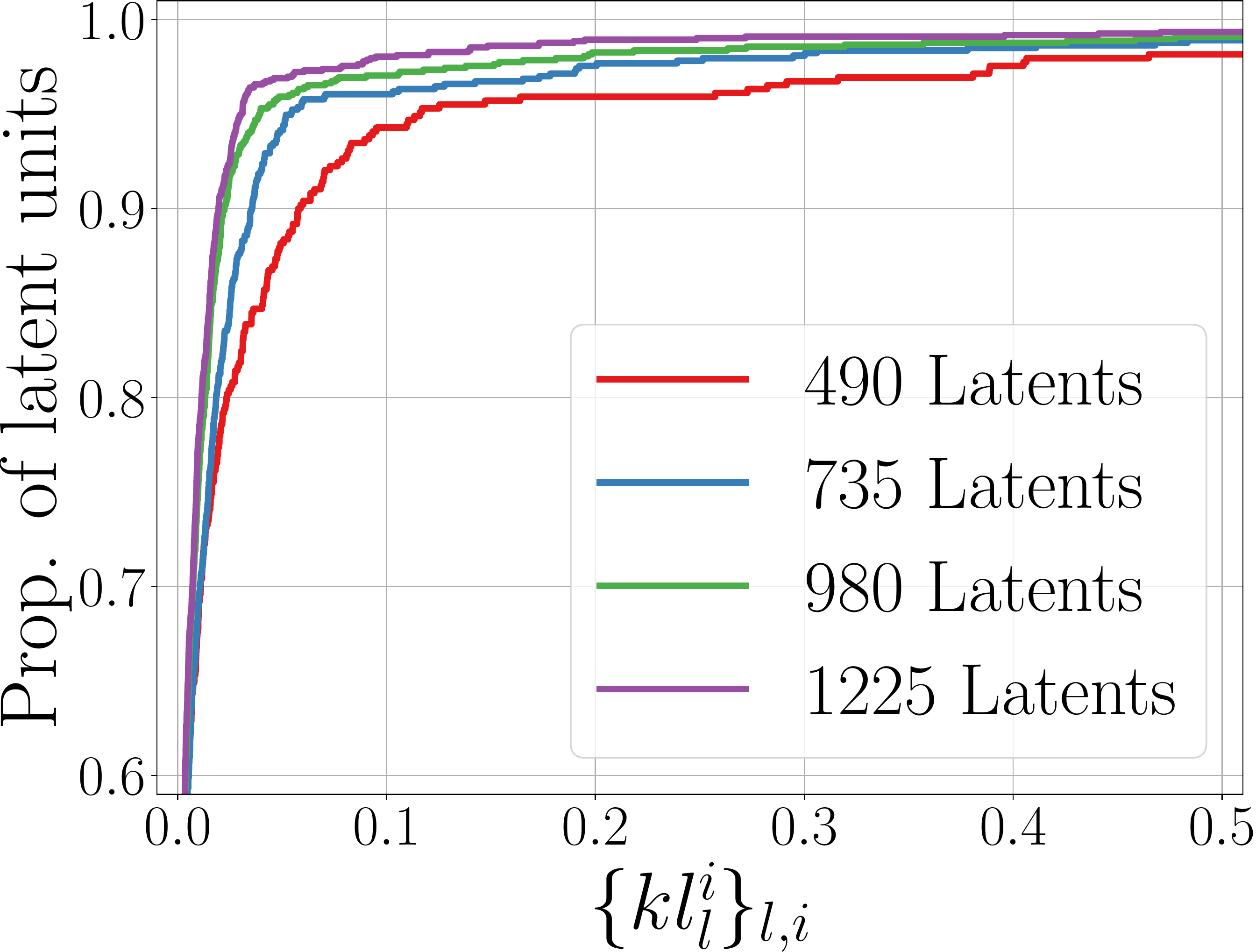}
    \vskip -10pt
    \captionof{figure}{The cumulative distribution function of the KL-divergence in VAEs with varying latent space sizes.}
    \label{fig:cum_kl_hist}
 \end{minipage}
 \vskip -12pt
 \end{table} 
 
In the following, we carry out an experiment to observe the posterior collapse in hierarchical VAE.
 We train four top-down hierarchical VAE models with different latent space sizes on the MNIST dataset. At the same time, we make sure that all the models have a similar number of parameters and try to keep the number of ResNet blocks the same. We vary the number of stochastic layers $L$ from 4 to 10. Note that the data space has a dimensionality of 784. 
We report the test NLL, Active Unit, and KL-divergence per latent variable for this experiment in Table \ref{tab:mnist_latent_size_au}. We also plot an empirical CDF of the latent variable's KLs in Figure \ref{fig:cum_kl_hist}.

The total number of latent units increases from 490 to 1225 in this experiment. However, all the models have no more than 40\% of active units. We also observe that AU and KL metrics decrease with the number of stochastic layers increasing. 
The cumulative histogram of KL-divergence (Eq. \ref{eq:kl_collapse_metric}) depicted in Figure \ref{fig:cum_kl_hist} shows that the models have close to 60\% of the latent variable with almost zero KL-divergence. 
This indicates that the deep hierarchical VAEs do not use the majority of the latent units. As a result, the common claim that the top-down hierarchical VAEs alleviate the problem of the posterior collapse \citep{maaloe2019biva} is not necessarily true as indicated by this experiment. It is true, though, that increasing the number of latents improves the performance (NLL). However, this is not an efficient way of utilizing the model since it disregards over $60\%$ of its latents.


\subsection{Latent variables non-identifiability and the posterior collapse in hierarchical VAEs}\label{subsec:post_collapse_and_non_ident}

\citet{wang2021posterior} prove that collapse of the true posterior in a one-level VAE takes place if and only if latent variables are \textit{non-identifiable}. 
A latent variable $\rvz$ is called non-identifiable~\citep{raue2009structural} if for a given set of parameter values $\theta^*$, the conditional likelihood does not depend on this latent variable. 
Namely, $p_{\theta^*}(\rvx|\rvz) = p_{\theta^*}(\rvx)$. Similarly, we say that latent variable $\rvz_l$ in hierarchical VAE is non-identifiable when $p_{\theta^*}(\rvx | \rvz_{1:L})=p_{\theta^*}(\rvx | \rvz_{-l})$.

We now establish the connection between posterior collapse (Eq.~\ref{eq:posterior_collapse_definition}) and non-identifiability in the following propositions. See Appendix~\ref{appx:theory_proof} for the proofs. 
\begin{propfirst}{prop:collapse_cond_indep} 
Consider a top-down hierarchical VAE introduced in Section \ref{sec:hierarchical_vae}. Then, for a given set of parameter values $\theta^*$, the posterior of the latent variable $\rvz_l$ collapses if and only if $\rvx$ and $\rvz_l$ are conditionally independent given ($\rvz_{l+1}, \dots, \rvz_L$).
\end{propfirst}

\begin{propfirst}{prop:cond_indep_to_identifiability}
Consider a top-down hierarchical VAE introduced in Section \ref{sec:hierarchical_vae}. If $\rvx$ and $\rvz_l$ are conditionally independent given ($\rvz_{l+1}, \dots, \rvz_L$), then the latent variable $\rvz_l$ is non-identifiable. However, if $\rvz_l$ is non-identifiable, it does not imply that it is conditionally independent with $\rvx$ given ($\rvz_{l+1}, \dots, \rvz_L$).
\end{propfirst}

To simplify the notation, let us split the latent variables of hierarchical VAEs into three groups:
\begin{align} \label{eq:three_groups}
    \underbrace{\rvz_1, \dots, \rvz_{l-1}}_{\rvz_A},  
    \rvz_l, 
    \underbrace{\rvz_{l+1}, \dots, \rvz_L}_{\rvz_C}.
\end{align}
We can do this for each $l \in {1, \dots, L}$, assuming that in the corner case of $l=1$, $\rvz_A$ is an empty set, and in the case of $l=L$, $\rvz_C$ is an empty set. Then, the content of the propositions~\ref{prop:collapse_cond_indep} and \ref{prop:cond_indep_to_identifiability} can be summarized in the following diagram: 
\begin{align*}
    \underbrace{p_{\theta^*}(\rvz_l | \rvz_C, \rvx) = p_{\theta^*}(\rvz_l | \rvz_C)}_{\text{\small{Posterior Collapse}}}\,\Leftrightarrow \, \underbrace{p_{\theta^*}(\rvx | \rvz_l, \rvz_C)=p_{\theta^*}(\rvx | \rvz_C)}_{\text{\small{Conditional Independence}}} \,\Rightarrow \,   \underbrace{p_{\theta^*}(\rvx | \rvz_A, \rvz_, \rvz_C)=p_{\theta^*}(\rvx | \rvz_A, \rvz_C)}_{\text{\small{Non-identifiability}}}.
\end{align*}

That being said, as opposed to the one-level VAE considered by \citep{wang2021posterior}, the non-identifiability of the latent variables in hierarchical VAEs does not necessarily cause the true posterior to collapse. 
Therefore, the solution, in which we define the likelihood function in a way that guarantees the latent variable identifiability might be too restrictive. 
One possible solution would be to utilize the method from \citep{wang2021posterior} to ensure that $\rvz_l$ and $\rvx$ are not conditionally independent given ($\rvz_{l+1}, \dots, \rvz_L$). 
However, one would need access to the distribution $p_{\theta^*}(\rvx|\rvz_l, \rvz_{l+1:L})$, which is intractable in the top-down hierarchical VAEs.

As a result, we employ an orthogonal approach by adding one more \textit{non-trainable} latent variable to a hierarchical VAE, which we call a \textit{context}. 
We show in Section \ref{subsec:context_posterior_collapse} that this method can break the link between conditional independence and posterior collapse without any restriction on the likelihood function.


\section{Hierarchical VAEs with non-trainable context} \label{sec:proposed_approach}

\subsection{Hierarchical VAEs with context} \label{subsec:context_definition}

In this work, we introduce a modified hierarchical VAE model, which is meant to increase the number of latent variables used by a deep hierarchical VAE while not harming performance. 
As we discuss in Sec.~\ref{subsec:post_collapse_and_non_ident}, posterior collapse happens if and only if there is a conditional independence between $\rvz_l$ and $\rvx$ given $\rvz_{>l}$. 
If this is the case, then the posterior distribution is proportional to the prior, namely,  $p_{\theta}(\rvz_l|\rvz_{l+1:L}, \rvx) \propto p_{\theta}(\rvz_l|\rvz_{l+1:L})$.
As a result, the latent variable $\rvz_l$ does not contain any information about the input $\rvx$. 
Note also that prior distribution is an object we can control since this is the distribution we parametrize directly by the neural network. 
This motivates us to introduce the context. 
We think of the context as a top-level latent variable that can be obtained from the input via a fixed, non-trainable transformation. 

\begin{wrapfigure}{r}{0.35\textwidth}
\vskip -15pt
    \centering
    \begin{tikzpicture}[node distance=.5cm]
 \node[obs, minimum size=0.75cm] (X) {$\rvx$};
 \node[det, minimum size=0.75cm, above=0.4cm of X] (d1) {$d_1$};
 \node[det, minimum size=0.75cm, above=0.4cm of d1] (d2) {$d_2$};
 \node[inner sep=0,minimum size=0.75cm, above=0.4cm of d2] (k) {}; 
 \node[latent, minimum size=0.75cm, right=0.5cm of k] (zL) {$\rvz_3$};
 \node[latent, minimum size=0.75cm, right=0.5cm of d1] (z1) {$\rvz_1$};
 \node[latent, minimum size=0.75cm, right=0.5cm of d2] (z2) {$\rvz_2$};
  \node[inner sep=0,minimum size=0.5cm, left=0.05cm of X] (k2) {}; 
  \node[inner sep=0,minimum size=0.5cm, right=0.05cm of zL] (k4) {}; 
  \edge[Latex-, thick] {d1} {X};
  \edge[-Latex, thick] {d1} {z1};
  \edge[Latex-, thick] {d2} {d1};
  \edge[-Latex, thick] {d2} {z2};
  \edge[-Latex, black!25!blue] {zL} {z2};
  \edge[-Latex, black!25!blue] {z2} {z1};
  \draw[-Latex, black!25!white] (X) -- (k2.center) |- node[above]{} (zL);
  \draw[-Latex, black!25!blue] (zL) -- (k4.center) |- (z1);
    \node[latent, minimum size=0.75cm, right=0.9cm of zL] (zL_gen) {$\rvz_3$};
    \node[latent, minimum size=0.75cm, below=0.4cm of zL_gen] (z2_gen) {$\rvz_2$};
    \node[latent, minimum size=0.75cm, below=0.4cm of z2_gen] (z1_gen) {$\rvz_1$};
    \node[obs, minimum size=0.75cm, below=0.4cm of z1_gen] (X_gen) {$\rvx$};
    \node[inner sep=0,minimum size=0.5cm, right=0.015cm of zL_gen] (k3) {}; 
    \node[inner sep=0,minimum size=0.5cm, right=0.25cm of zL_gen] (k5) {}; 
    \node[inner sep=0,minimum size=0.5cm, left=0.05cm of z2_gen] (k6) {}; 
    \edge[-Latex, black!25!blue] {zL_gen} {z2_gen};
    \edge[-Latex, black!25!blue] {z2_gen} {z1_gen};
    \draw[-Latex, black!25!blue] (zL_gen) -- (k3.center) |- (z1_gen);
    \draw[-Latex] (zL_gen) -- (k5.center) |- (X_gen);
    \draw[-Latex] (z2_gen) -- (k6.center) |- (X_gen);
    \edge[-Latex, thick] {z1_gen} {X_gen};

\end{tikzpicture}
    \caption{Graphical model of the top-down hierarchical VAE with two latent variables and the context $\rvz_3$. The inference model (left) and the generative model (right) share the top-down path (blue). The grey arrow represents a non-trainable transformation.}
    \label{fig:dct_vae_graph_model}
\vskip -20pt
\end{wrapfigure}
Let us consider the top latent variable $\rvz_L$ to be given by a non-learnable transformation of the input $\rvx$, namely, $\rvz_L = f(\rvx)$. 
We require context $\rvz_L$ to be a much simpler object than the initial object $\rvx$.  
That is, we want the dimensionality of $\rvz_L \in \mathbb{R}^{M_L}$ to be smaller than the dimensionality of $\rvx \in \mathcal{X}^D$, $M_L \ll D$. 
At the same time, we want the context to be a reasonable representation of $\rvx$.
We can think of the context as a \textit{compressed} representation of the input data, e.g., in the simplest case, it could be a downsampled version of an image (see Appendix \ref{appx:context_down} for details).
We discuss another way of constructing the context in Section~\ref{subsec:dct_context}.

The graphical model of the VAE with the context is depicted in Figure~\ref{fig:dct_vae_graph_model}. 
We use the top-down VDVAE architecture \citep{child2020very} and extend this model with a deterministic, non-trainable function to create latent variable $\rvz_L$ (the context). Context $\rvz_L$ is produced from the observation $\rvx$ and further used to condition all other latent variables in both inference and generative models.  
We provide a mode details on the architecture in Appendix \ref{appendix:schema} (Figure \ref{fig:model_schema})

\subsection{Training VAE with the context}

We assume that both $\rvx$ and $\rvz_L$ are discrete random variables. 
Furthermore, we assume that the variational posterior of the context is Kronecker's delta function $q(\rvz_L|\rvx) = \delta(\rvz_L - f(\rvx))$. 
As we depict in Figure~\ref{fig:dct_vae_graph_model}, the generative model is conditioned on the context latent variable $\rvz_L$ at each step. 
To sample unconditionally, we define a context prior distribution $p_{\gamma}(\rvz_L)$, which is trained simultaneously with the whole VAE model via the ELBO objective. 
Following \citep{vahdat2021score, wehenkel2021diffusion}, we propose to use a diffusion-based generative model \citep{ho2020denoising} as the prior. 
Since the context is a less complex object, we assume that it is enough to use a model much smaller compared to the VAE itself. 
We provide details on diffusion models in Appendix \ref{appendix:ddgm_theory}. 
The diffusion-based model provides a lower bound on the log density of the prior distribution $\mathcal{L}(\gamma, \rvz_L) \leq \ln p_{\gamma}(\rvz_L)$, which together with VAE objective~\ref{eq:vae_elbo} results in the following objective:

\small{
\begin{align}\label{eq:our_elbo}
     &\E_{\Enc{\rvz_{1:L}}{\rvx}} \left[\ln \Dec{\rvx}{\rvz_{1:L}} \right] + \E_{q(\rvz_L|\rvx)}\mathcal{L}(\gamma, \rvz_L)
     - 
     \sum_{l=1}^{L-1} \E_{q_{\phi}(\rvz_{l+1:L}|\rvx)}\KL{q_{\phi}(\rvz_{l}|\rvz_{l+1:L}, \rvx)}{p_{\theta}(\rvz_{l} | \rvz_{l+1:L})}  . \notag
\end{align}
\small}
\normalsize
 
\subsection{The posterior collapse for VAEs with the context} \label{subsec:context_posterior_collapse}
We claim that the introduction of the context changes the prior distributions, which results in the posterior collapse having less effect on the model. 
First, since $\rvz_L = f(\rvx)$, we guarantee that the top latent variable will not collapse.
We now need to fit the prior to the aggregated posterior $q(\rvz_L) = \sum_{\rvx} \delta(\rvz_L - f(\rvx)) q(\rvx)$, not the other way around. 
As a result, this prior contains information about the data points $\rvx$ by definition.
Second, let us assume that $\rvz_{l}$ and $\rvx$ are conditionally independent for given parameter values $\theta^*$: $p_{\theta}(\rvx |\rvz_{l}, \rvz_{l+1:L}) = p_{\theta}(\rvx| \rvz_{l+1:L})$. 
Then, from the Proposition \ref{prop:collapse_cond_indep} the posterior is proportional to the prior:
$p_{\theta}(\rvz_{l} | \rvz_{l+1:L}, \rvx)   \propto  p_{\theta}(\rvz_{l}| \rvz_{l+1:L}).$
However, since $f(\rvx) = \rvz_L \in \rvz_{l+1:L}$, we still have information about $\rvx$ preserved in the posterior:
\begin{align}
  p_{\theta}(\rvz_{l} | \rvz_{l+1:L}, \rvx)   \propto p_{\theta}(\rvz_{l}| \rvz_{l+1:L-1}, f(\rvx)).
\end{align}
This way, the presence of posterior collapse does not necessarily lead to uninformative latent codes.

\subsection{A DCT-based context}\label{subsec:dct_context}

We suggest to think of the context as of \textit{compressed} representation of the input data (Sec.~\ref{subsec:context_definition}). 
We expect it to be lower-dimensional compared to the data itself while preserving crucial information. 
In other words, we may say that context does not contain any high-frequency details of the signal of interest while preserving a more general pattern.  
 To this end, we propose to use the Discrete Cosine Transform\footnote{In this work, we consider the most widely used type-II DCT.} (DCT) to create the context.
 DCT\citep{ahmed1974discrete} is widely used in signal processing for image, video, and audio data, i.e., it is a part of the JPEG standard \citep{pennebaker1992jpeg}. 
 DCT is a linear transformation that decomposes a discrete signal on a basis consisting of cosine functions of different frequencies.

 Let us consider a signal as a $3D$ tensor $\rvx \in \mathcal{X}^{\text{Ch} \times D \times D}$. 
 Then DCT for a single channel, $\rvx_{i}$, is defined as follows: $\rvz_{DCT, i} = \rmC \rvx_{i} \rmC^{\top}$, 
where for all pairs $(k=0,n)$: $\rmC_{k,n} = \sqrt{\frac{1}{D}}$, 
and for all pairs $(k,n)$ such that $k>0$: $\rmC_{k,n} = \sqrt{\frac{2}{D}} \cos \left(\frac{\pi}{D}\left(n+\frac{1}{2}\right) k\right)$. 
A helpful property of the DCT is that it is an \textit{invertible} transformation. Therefore, it contains all the information about the input. 
However, for our approach, we want the context to be lower-dimensional compared to the input dimensionality. 
Therefore, we propose to remove high-frequency components from the signal. 
Assume that each channel of $\rvx$ is $D\times D$. We select the desired size of the context $d < D$ and remove (crop) $D-d$ bottom rows and right-most columns for each channel in the frequency domain. 
Finally, we perform normalization using matrix $\rmS$, which contains the maximal absolute value of each frequency. 
We calculate this matrix using all the training data: $\rmS = \max_{\rvx \in \mathcal{D}_{\text{train}}} |\text{DCT}(\rvx)|$. As a result, we get latent variables whose values are in $[-1, 1]$. In the last step, we round all values to a given precision such that after multiplying the latents by $\rmS$ we get integers, thus, we get discrete variables. We call this the \textit{quantization} step. 
Algorithm~\ref{alg:context_forward} describes context computation from the given input $\rvx$.

\begin{minipage}[t]{0.46\textwidth}
\vskip -15pt
\begin{algorithm}[H]
\caption{Create a DCT-based context}
\label{alg:context_forward}
\begin{algorithmic}
        \State \hskip-3mm \textbf{Input}: $\rvx, \rmS, d$
    \State $\rvz_{\text{DCT}} = \text{DCT}(\rvx)$
        \State $\rvz_{\text{DCT}} = \text{Crop}(\rvz_{\text{DCT}}, d)$ 
        \State $\rvz_{\text{DCT}} = \frac{\rvz_{\text{DCT}}}{\rmS} $  
        \State $\rvz_{\text{DCT}} = \text{quantize}(\rvz_{\text{DCT}})$
        \State  \hskip-3mm \textbf{Return}: $\rvz_{\text{DCT}}$
\end{algorithmic}
\end{algorithm}
\end{minipage}
\hfill
\begin{minipage}[t]{0.46\textwidth}
\vskip -15pt
\begin{algorithm}[H]
\caption{Decode the DCT-based context.}
\label{alg:context_backward}
\begin{algorithmic} 
        \State \hskip-3mm \textbf{Input}: $\rvz_{\text{DCT}}, \rmS, D$
        \State $\rvz_{\text{DCT}} = \rvz_{\text{DCT}} \cdot \rmS$
        \State $\rvz_{\text{DCT}} = \text{zero\_pad}(\rvz_{\text{DCT}}, D-d)$
        \State $\tilde{\rvx}_{context} = \text{iDCT}(\rvz_{\text{DCT}})$ 
        \State \hskip-3mm \textbf{Return}: $\tilde{\rvx}_{context}$
\end{algorithmic}
\end{algorithm}
\end{minipage}

Due to cropping and quantization operations, the context computation is not invertible anymore. 
However, we can still go back from the frequency to the local domain. 
First, we start by multiplying by the normalization matrix $\rmS$. 
Afterwards, we pad each channel with zeros, so that the size increases from $d\times d$ to $D\times D$. 
Lastly, we apply the inverse of the Discrete Cosine Transform (iDCT). 
We describe this procedure in Algorithm~\ref{alg:context_backward}.  
We refer to our top-down hierarchical VAE with a DCT-based context as DCT-VAE.



\section{Experiments}
We evaluate DCT-VAE on several commonly used image datasets, namely, MNIST, OMNIGLOT, and CIFAR10. We provide the full set of hyperparameters in Appendix \ref{appendix:hyperparams}.
We designed the experiments to validate the following hypotheses:

$1)$ \textit{Adding the DCT-based context into hierarchical VAE does not harm the performance (as measured by negative loglikelihood)} (sec. \ref{sect:exp_image_generations}).\\
$2)$ \textit{DCT-VAE have more active units / higher KL values} (sec. \ref{sect:posterior_collapse}).\\
$3)$ \textit{Latent variables of very deep DCT-VAE carry more information about the input data} (sec. \ref{sect:compression}).

In all the experiments, we implement two models: A baseline Very Deep VAE model without any context (denoted by VDVAE) \citep{child2020very}, and our approach (DCT-VAE) that is a VDVAE with a DCT-based context on top. 
We keep both architectures almost identical, keeping the same number of channels, resnet blocks, and latent space sizes.
In other words, the only difference in the architecture is the presence of the context in DCT-VAE. 

\subsection{Image generation benchmarks}\label{sect:exp_image_generations}
\paragraph{Binary images}
We start with the experiments on binary images: MNIST and OMNIGLOT, for which we use dynamic binarization. In Figure \ref{fig:mnist_results_bar}, we report the results of an ablation study where we test various context sizes and two contexts: downsampling and DCT. We observe that DCT-VAE (green) outperforms the VDVAE in all the experiments (the orange horizontal line). However, if we choose downsampling as a context instead of the DCT, the performance of the model drops significantly for larger context sizes (blue bars). The reason for that comes from the fact that it becomes harder to fit the prior to the aggregated posterior. Interestingly, it seems there is a \textit{sweet spot} for the context size of the DCT-VAE at around $5\%$. Since DCT always performs better than downsampling, we use it in all the experiments from now on.
Comparing DCT-VAE to various best-performing VAEs, it turns out that our approach not only does not harm performance but also achieves state-of-the-art performance on both datasets, see Table \ref{tab:main_results}. Importantly, the introduction of the context gives a significant improvement over the same architecture of the VDVAE.  

 \begin{table}
    \begin{minipage}[c]{0.65\textwidth}
        \centering
        \caption{The test performance (NLL) on MNIST and OMNIGLOT datasets and the number of stochastic layers ($L$).}
        \label{tab:main_results_mnists}
        \vskip 5pt
        \begin{tabular}{l||lcc}
            \toprule
            \multirow{2}{*}{\textsc{Model}} & \multirow{2}{*}{\textsc{L}} & \footnotesize{\textsc{MNIST}} & \footnotesize{\textsc{OMNIGLOT}}\\ 
                & &  \multicolumn{2}{c}{$ - \log p(\rvx) \leq \, \downarrow $} \\
            \midrule
            \textbf{DCT-VAE} \footnotesize{(ours)} & 8 
                    & \textbf{76.62} & \textbf{86.11} \\
            \textbf{Donwsample-VAE} \footnotesize{(ours)} & 8 
                    & 77.52 & 87.69 \\
            Small VDVAE  & \multirow{2}{*}{8} 
                    & \multirow{2}{*}{78.27} & \multirow{2}{*}{88.14}           \\
                     (our implementation)\\
            Attentive VAE   & \multirow{2}{*}{15}
                    & \multirow{2}{*}{77.63} & \multirow{2}{*}{89.50}                       \\
                    \footnotesize{\citep{apostolopoulou2021deep}}  \\
            CR-NVAE              & \multirow{2}{*}{15}
                    & \multirow{2}{*}{76.93} & \multirow{2}{*}{---}  \\
                    \small{\citep{sinha2021consistency}} \\
            OU-VAE  & \multirow{2}{*}{5}
                    & \multirow{2}{*}{81.10} & \multirow{2}{*}{96.08}         \\
                    \small{\citep{pervez21a}} \\
            NVAE & \multirow{2}{*}{15}
                    & \multirow{2}{*}{78.01} & \multirow{2}{*}{---}   \\
                     \small{\citep{vahdat2020nvae}} \\
            BIVA\small{\citep{maaloe2019biva}}   & 6
                    & 78.41 & 91.34       \\
            LVAE  & \multirow{2}{*}{5}
                    & \multirow{2}{*}{81.74} & \multirow{2}{*}{102.11}      \\
                    \small{\citep{sonderby2016ladder}} \\
            IAF-VAE & \multirow{2}{*}{---}
                    & \multirow{2}{*}{79.10}  &   \multirow{2}{*}{---}             \\
                     \small{\citep{kingma2016improved}} \\
            \bottomrule
        \end{tabular}
        \vskip -20pt
    \end{minipage}\hfill 
    \begin{minipage}[c]{0.35\textwidth}
        \begin{tabular}{c}
        \includegraphics[width=0.85\linewidth]{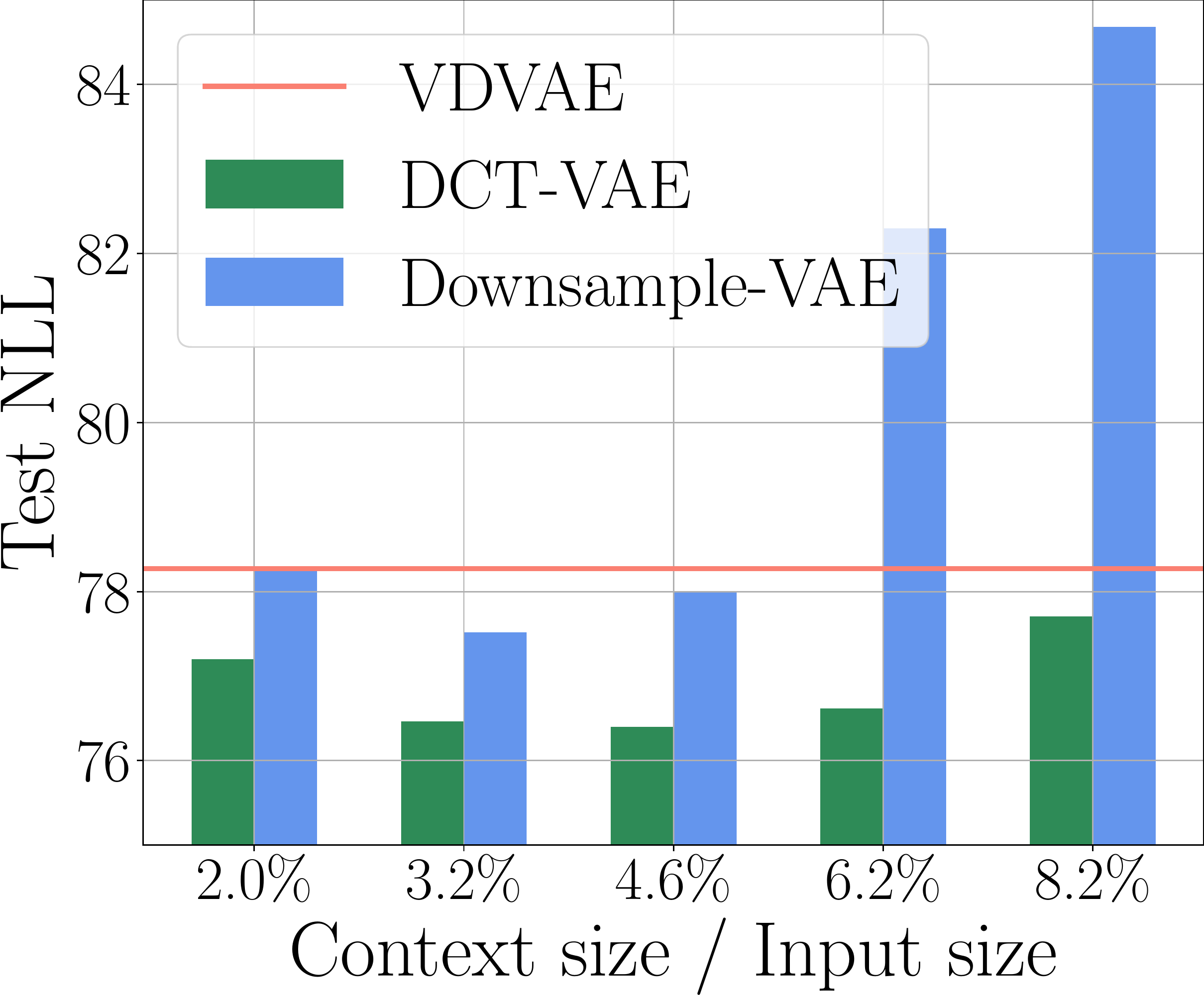} \\
        (a) \footnotesize{\textsc{MNIST}} \\
        \includegraphics[width=0.85\linewidth]{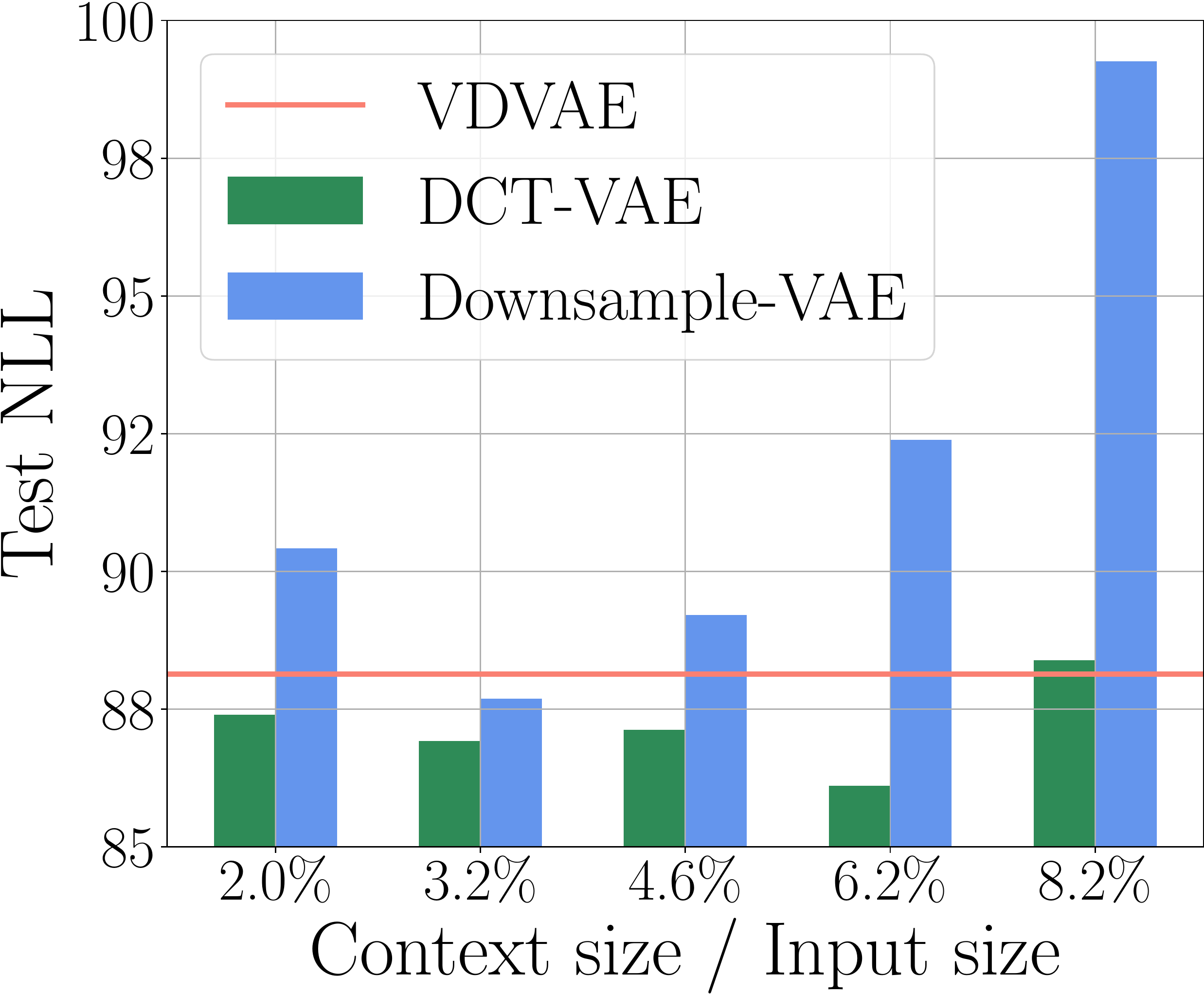} \\
        (b) \footnotesize{\textsc{OMNIGLOT}} \\
        \end{tabular}
        \captionof{figure}{NLL results for MNIST and OMIGLOT for different context types and sizes.}
        \label{fig:mnist_results_bar}
        \vskip -20pt
    \end{minipage}
 \end{table} 
 

\paragraph{Natural Images}

\begin{wraptable}{r}{0.6\textwidth}
    \centering
    \vskip -20pt
    \caption{The test performance (BPD) on the CIFAR10 dataset, the total number of trainable parameters (Size), the number of stochastic layers ($L$).}
    \vskip 5pt
    \label{tab:main_results}
    \begin{tabular}{l||rlc}
        \toprule
        \textsc{Model}  & 
         \textsc{Size} & \textsc{L} & \small{\textsc{bits/dim $ \leq \, \downarrow $}} \\
        \midrule
        \textbf{DCT-VAE (ours)}& 22M & 29 & 3.26  \\
        Small VDVAE    & \multirow{2}{*}{21M} & \multirow{2}{*}{29} & \multirow{2}{*}{3.28}  \\
        (our implementation) \\
        OU-VAE  &  \multirow{2}{*}{10M} & \multirow{2}{*}{3} & \multirow{2}{*}{3.39}  \\
        \small{\citep{pervez21a}}\\
        Residual flows& \multirow{2}{*}{25M} & \multirow{2}{*}{1}
                & \multirow{2}{*}{3.28}  \\
                 \small{\citep{perugachi2021invertible}}\\
        i-DenseNet flows & \multirow{2}{*}{25M} & \multirow{2}{*}{1}
                & \multirow{2}{*}{3.25}  \\
                \small{\citep{perugachi2021invertible}}\\
        \bottomrule
    \end{tabular}
    \vskip -5pt
\end{wraptable}
We perform experiments on natural images to test the method's performance on a more challenging task.
We use the CIFAR10 dataset, which is a common benchmark in VAE literature. 

We note that the best-performing VAEs (e.g., VDVAE, NVAE) on this dataset are very large and require substantial computational resources to train which we do not have access to. Instead, we train a small-size VDVAE and provide results of other generative models of comparable sizes in Table \ref{tab:main_results}. We report the complete comparison (including large models) in Appendix~\ref{appx:cifar_full}.

We observe that our approach works on par with the generative models that have comparable sizes (OU-VAE, Residual Flows, GLOW), and, most importantly, it has a similar (in fact, slightly better) BPD to our implementation of the VDVAE of a similar size.


\subsection{Posterior collapse}\label{sect:posterior_collapse}

In this section, we analyze the latent space of the DCT-VAE and VDVAE trained on different datasets from the posterior collapse point of view. We report the number of active units and KL-divergence on the test dataset in Table \ref{tab:au_results}.
We also show the total latent space size and context size. 

We observe that the number of active units increases significantly when the context is introduced to the model. 
Furthermore, this increase is much higher than the size of the context itself, meaning that it helps to increase the latent space utilization in general.
However, there are still a lot of unused latent variables. For example, on the CIFAR10 dataset, the proportion of active units increases from $7\%$ to $11\%$. It means that even though deeper models obtain better NLL, there is still a significant waste of the model's capacity.
Similarly to the AU metric, the higher KL-divergence of the DCT-VAE compared to the VDVAE with no context indicates that the DCT-based context helps to \textit{push} more information to other layers. In conclusion, we observe the improved utilization of latent space in terms of both metrics.

\begin{table}[t] 
\centering
    \caption{The absolute and the relative number of active units for VAEs and DCT-VAEs evaluated on the test datasets of MNIST, OMNIGLOT, and Cifar10.} 
    \label{tab:au_results}
    \vskip 5pt
    \begin{tabular}{lccccc}
        \toprule
          & \small{\textsc{Latent}}& \small{\textsc{Context}} & \small{\textsc{AU}$\uparrow$} & \small{\textsc{AU}$\uparrow$} &  \small{\textsc{KL}$\uparrow$}\\
     &\small{\textsc{Space}} & \small{\textsc{Size}}& \footnotesize{(Absolute)} &  \footnotesize{(\% of latents)}& \footnotesize{(per latent unit)} \\
     &&&&& \footnotesize{$\times 10e-3$}\\
        \midrule 
         &\multicolumn{5}{c}{\textsc{MNIST}} \\ \cmidrule(r){2-6}
          VDVAE & 980 & 0 & 336 & 34.4\% & 22.9 \footnotesize{(1.4)}\\
           DCT-VAE & 967 & 36 & \textbf{405} & \textbf{41.9}\% & \textbf{25.9} \footnotesize{(0.8)}\\ \cmidrule(r){1-6}
          & \multicolumn{5}{c}{\textsc{OMNIGLOT}} \\ \cmidrule(r){2-6}
         VDVAE & 980 & 0 & 494 & 50.4\% & 35.1 \footnotesize{(0.8)}\\
          DCT-VAE & 980 & 49 & \textbf{593} & \textbf{60.5}\% & \textbf{36.5} \footnotesize{(0.8)}\\ \cmidrule(r){1-6}
           &\multicolumn{5}{c}{\textsc{CIFAR10}} \\ \cmidrule(r){2-6}
          VDVAE & 105K& 0 & 7.5K & 7.1\% & 47.6 \footnotesize{(2.1)}\\
          DCT-VAE & 105K & 108 & \textbf{11.3K} & \textbf{10.8}\% & \textbf{51.6} \footnotesize{(2.0)}\\
        \bottomrule
    \end{tabular}
\end{table}

\subsection{Data information in latent variables}\label{sect:compression}
Many of the state-of-the-art models have a lot of stochastic layers (e.g., 45 for CIFAR10 \citep{child2020very}). 
Therefore, it is likely that the information about the $\rvx$ could be completely disregarded by the latent variables further away from the input. 
In this section, we explore how much information about the corresponding data points the top latent codes contain. 
For this purpose, we consider the reconstruction performance and compression. We examine VDVAE and DCT-VAE with 29 stochastic layers trained and tested on the CIFAR10 dataset in both experiments. 

\subsubsection{Reconstruction capabilities of DCT-VAE} 
We compute Multi-Scale Structural Similarity Index Measure ($\mathrm{MSSSIM}$) \citep{wang2003multiscale} for the test data and its reconstruction obtained using only part of the latent variables from the variational posterior.
That is, for each $m \in \{1,\dots, L\}$ we obtain a reconstruction $\Tilde{\rvx}^m$ using $m$ latent variables from the variational posterior and by sampling the rest $L-m$ latent variables from the prior, namely:  
\begin{align}
  \Tilde{\rvx}^m \sim& \Dec{\cdot}{\rvz_{1:L}}  \prod_{l=1}^{L-m}p_{\theta}(\rvz_l | \rvz_{l+1:L}) 
  \prod_{l=L-m+1}^{L}\Enc{\rvz_l}{\rvz_{l+1:L}, \rvx}.
\end{align}
We present the results of this experiment in Figure \ref{fig:gen_rec}. We observe that in VDVAE the top latent layers carry very little to no information about the real data point $\rvx$, which continues up to the $5^{\text{th}}$ layer from the top. Then, the reconstructions become reasonable (between the $5^{\text{th}}$ and the $10^{\text{th}}$ layer values of $\mathrm{MSSSIM}$ increases from 0.6 to 0.8). In the case of DCT-VAE, using only one layer (i.e., context) gives already reasonable reconstructions ($\mathrm{MSSSIM}$ above 0.8).

\begin{table}[t]
\begin{minipage}[c]{0.388\textwidth}
\includegraphics[width=0.85\linewidth]{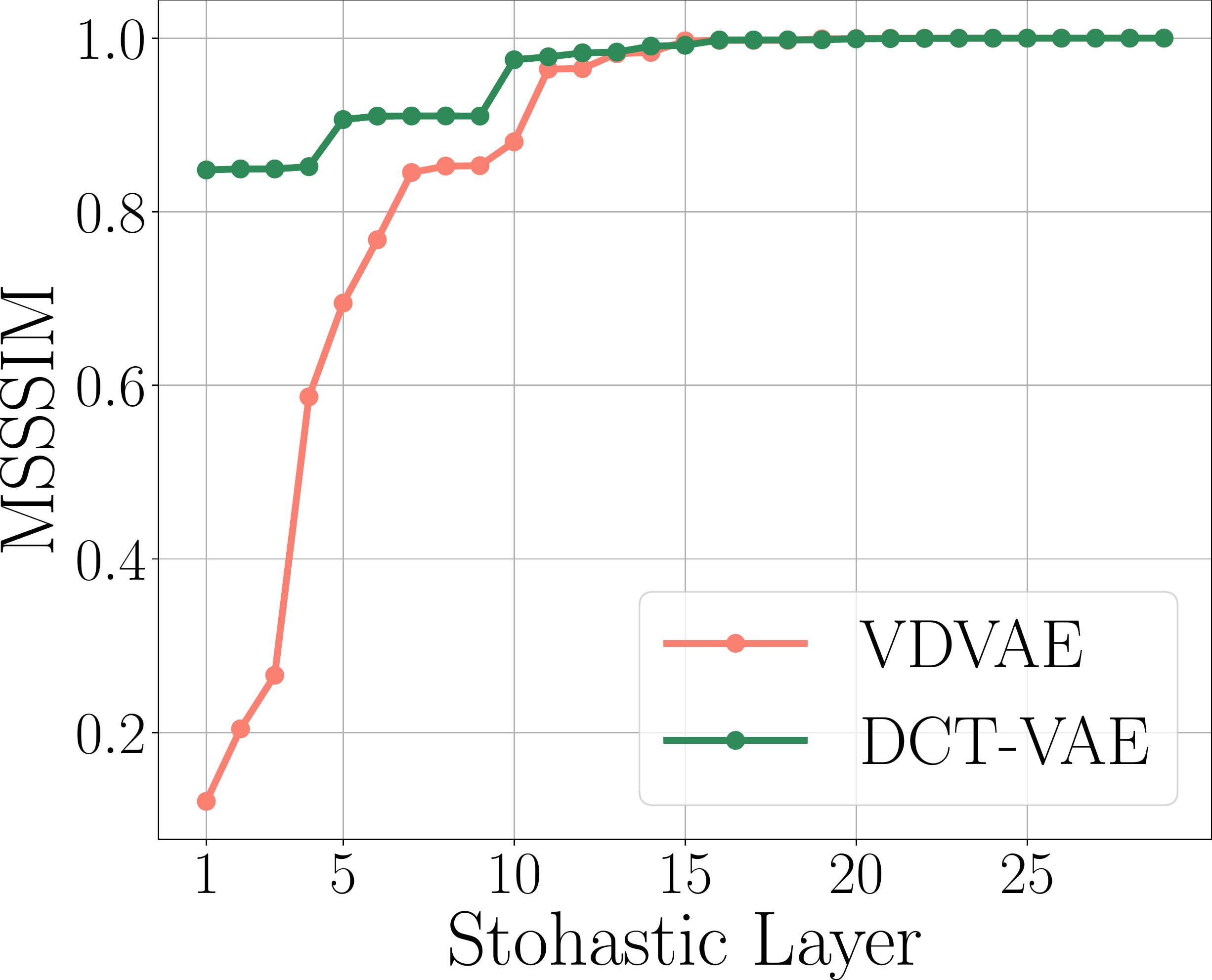} 
\vskip -10pt
 \captionof{figure}{The reconstruction measured by the MSSSIM ($\uparrow$) on the CIFAR10 test set for a varying number of latent variables sampled from the encoder. }
    \label{fig:gen_rec}
\end{minipage}\hfill
\begin{minipage}[c]{0.60\textwidth}
\includegraphics[width=1.0\linewidth]{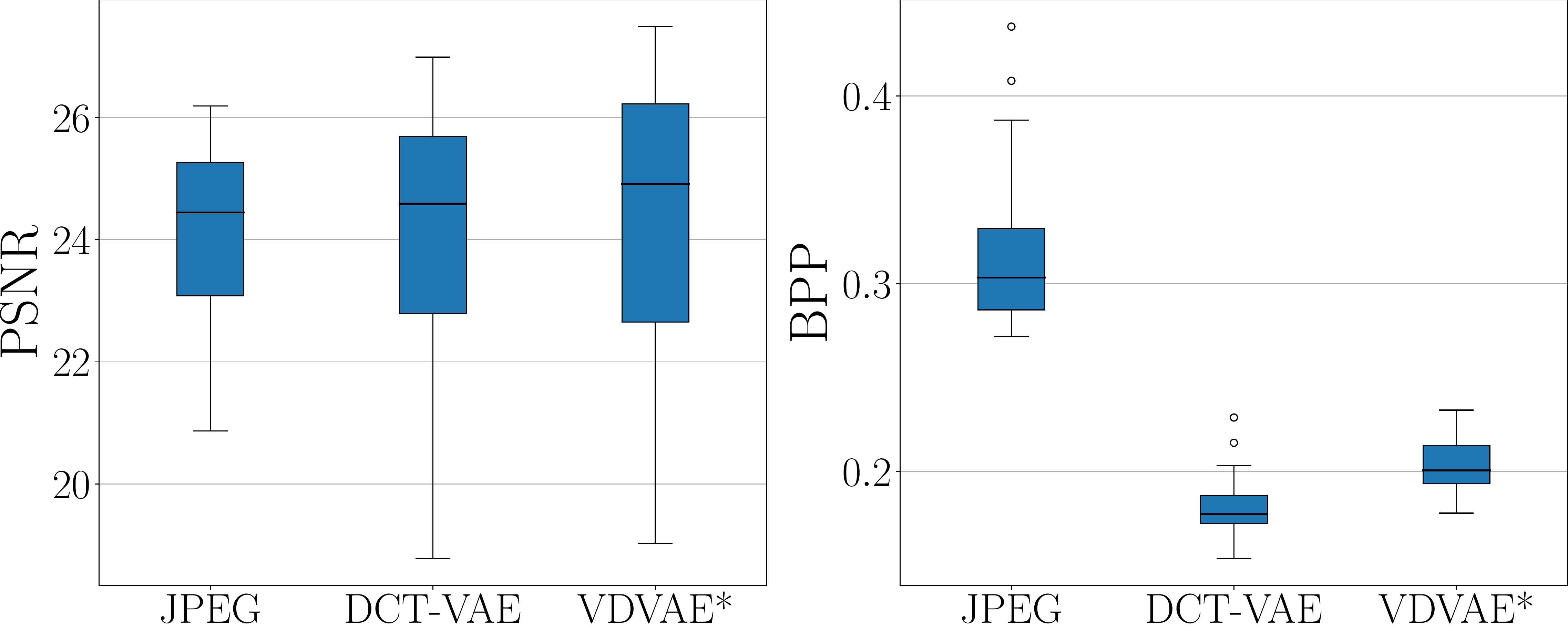} 
\vskip -2pt
 \captionof{figure}{Compression result on KODAK dataset. We use discrete context only to compress images with DCT-VAE. We report the BPP of JPEG and VDVAE that corresponds to the same reconstruction quality. }
    \label{fig:compression_1_latent}
\end{minipage}
\end{table}

\begin{figure}[t]
    \begin{tabular}{>{\centering}p{0.3\textwidth}>{\centering}p{0.3\textwidth}>{\centering\arraybackslash}p{0.3\textwidth}}
        \multicolumn{3}{c}{\includegraphics[width=0.95\textwidth]{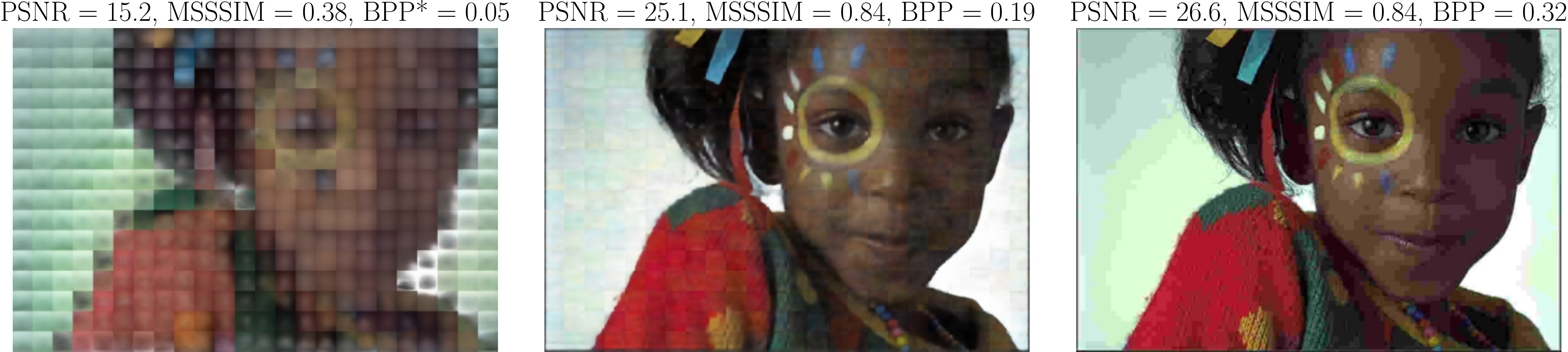} } \\
        (a) VDVAE & (b) DCT-VAE  &(c) JPEG \\
    \end{tabular}
    \vskip -10pt
    \caption{Examples of the decompressed images. We use  (a) 2 top latent variables of VDVAE to reconstruct the image, (b) only the context of DCT-VAE, and (c) we choose JPEG compression to have a similar PSNR value to DCT-VAE.}
    \label{fig:cifar_kodak_example_short}
\end{figure}

\subsubsection{Image compression with DCT-VAE} 
To find out how much information about the data is preserved in the top latent variable, we conduct an experiment
in which we use the baseline VDVAE and the DCT-VAE pretrained on CIFAR10 for compression. 
We use the KODAK dataset, which is a standard compression benchmark containing 24 images with resolution $512 \times 768$. 
Since CIFAR10 images are $32 \times 32$, we independently encode patches of KODAK images. 
We then reconstruct each patch using only the context latent variable, while the rest of the latent variables are sampled from the prior. 
We combine these patches to obtain final reconstructions and measure reconstruction error (PSNR). We use JPEG as a baseline.

Results are provided in Figure \ref{fig:compression_1_latent}. 
We select the compression rates that result in comparable PSNR values. 
We report KL-divergence converted to bits-per-pixel as a theoretical compression rate. All the latent variables (except for the context in DCT-VAE) are continuous.
We provide an example of the KODAK image after compression in Figure~ \ref{fig:cifar_kodak_example_short}. We also plot examples of the reconstructed images in the Appendix Figure~\ref{fig:cifar_kodak_example_extended}. Interestingly, DCT-VAE is capable of obtaining much better \textsc{BPP} than two other baselines while keeping the same PSNR. This indicates the usefulness of context.

\section{Conclusion}

In this paper, we discuss the issue of posterior collapse in top-down hierarchical VAEs. 
We show theoretically and empirically that this problem exists. 
As a solution, we propose to introduce deterministic, discrete and non-trainable transformations to calculate the top latent variables, e.g., DCT. 
The resulting model, DCT-VAE, seems to give more robust latent variables that carry more information about data (e.g., the compression experiment).


\bibliography{iclr2024_conference}

\begin{thebibliography}{41}
\providecommand{\natexlab}[1]{#1}
\providecommand{\url}[1]{\texttt{#1}}
\expandafter\ifx\csname urlstyle\endcsname\relax
  \providecommand{\doi}[1]{doi: #1}\else
  \providecommand{\doi}{doi: \begingroup \urlstyle{rm}\Url}\fi

\bibitem[Ahmed et~al.(1974)Ahmed, Natarajan, and Rao]{ahmed1974discrete}
Nasir Ahmed, T~Natarajan, and Kamisetty~R Rao.
\newblock Discrete cosine transform.
\newblock \emph{IEEE Transactions on Computers}, 100\penalty0 (1):\penalty0 90--93, 1974.

\bibitem[Alemi et~al.(2018)Alemi, Poole, Fischer, Dillon, Saurous, and Murphy]{alemi2018fixing}
Alexander Alemi, Ben Poole, Ian Fischer, Joshua Dillon, Rif~A Saurous, and Kevin Murphy.
\newblock Fixing a broken elbo.
\newblock In \emph{ICML}, 2018.

\bibitem[Apostolopoulou et~al.(2022)Apostolopoulou, Char, Rosenfeld, and Dubrawski]{apostolopoulou2021deep}
Ifigeneia Apostolopoulou, Ian Char, Elan Rosenfeld, and Artur Dubrawski.
\newblock Deep attentive variational inference.
\newblock In \emph{ICLR}, 2022.

\bibitem[Bowman et~al.(2016)Bowman, Vilnis, Vinyals, Dai, Jozefowicz, and Bengio]{bowman2015generating}
Samuel Bowman, Luke Vilnis, Oriol Vinyals, Andrew Dai, Rafal Jozefowicz, and Samy Bengio.
\newblock Generating sentences from a continuous space.
\newblock In \emph{SIGNLL}, 2016.

\bibitem[Burda et~al.(2015)Burda, Grosse, and Salakhutdinov]{burda2015importance}
Yuri Burda, Roger Grosse, and Ruslan Salakhutdinov.
\newblock Importance weighted autoencoders.
\newblock \emph{arXiv}, 2015.

\bibitem[Child(2021)]{child2020very}
Rewon Child.
\newblock Very deep vaes generalize autoregressive models and can outperform them on images.
\newblock In \emph{ICLR}, 2021.

\bibitem[Dhariwal \& Nichol(2021)Dhariwal and Nichol]{dhariwal2021diffusion}
Prafulla Dhariwal and Alexander Nichol.
\newblock Diffusion models beat gans on image synthesis.
\newblock \emph{NeurIPS}, 2021.

\bibitem[Dieng et~al.(2019)Dieng, Kim, Rush, and Blei]{dieng2019avoiding}
Adji~B Dieng, Yoon Kim, Alexander~M Rush, and David~M Blei.
\newblock Avoiding latent variable collapse with generative skip models.
\newblock In \emph{AISTATS}, 2019.

\bibitem[Fu et~al.(2019)Fu, Li, Liu, Gao, Celikyilmaz, and Carin]{fu2019cyclical}
Hao Fu, Chunyuan Li, Xiaodong Liu, Jianfeng Gao, Asli Celikyilmaz, and Lawrence Carin.
\newblock Cyclical annealing schedule: A simple approach to mitigating kl vanishing.
\newblock \emph{arXiv}, 2019.

\bibitem[G{\'o}mez-Bombarelli et~al.(2018)G{\'o}mez-Bombarelli, Wei, Duvenaud, Hern{\'a}ndez-Lobato, S{\'a}nchez-Lengeling, Sheberla, Aguilera-Iparraguirre, Hirzel, Adams, and Aspuru-Guzik]{gomez2018automatic}
Rafael G{\'o}mez-Bombarelli, Jennifer~N Wei, David Duvenaud, Jos{\'e}~Miguel Hern{\'a}ndez-Lobato, Benjam{\'\i}n S{\'a}nchez-Lengeling, Dennis Sheberla, Jorge Aguilera-Iparraguirre, Timothy~D Hirzel, Ryan~P Adams, and Al{\'a}n Aspuru-Guzik.
\newblock Automatic chemical design using a data-driven continuous representation of molecules.
\newblock \emph{ACS central science}, 2018.

\bibitem[Havrylov \& Titov(2020)Havrylov and Titov]{havrylov2020preventing}
Serhii Havrylov and Ivan Titov.
\newblock Preventing posterior collapse with levenshtein variational autoencoder.
\newblock \emph{arXiv}, 2020.

\bibitem[He et~al.(2019)He, Spokoyny, Neubig, and Berg-Kirkpatrick]{he2019lagging}
Junxian He, Daniel Spokoyny, Graham Neubig, and Taylor Berg-Kirkpatrick.
\newblock Lagging inference networks and posterior collapse in variational autoencoders.
\newblock In \emph{ICLR}, 2019.

\bibitem[Ho et~al.(2020)Ho, Jain, and Abbeel]{ho2020denoising}
Jonathan Ho, Ajay Jain, and Pieter Abbeel.
\newblock Denoising diffusion probabilistic models.
\newblock \emph{NeurIPS}, 2020.

\bibitem[Huang et~al.(2021)Huang, Lim, and Courville]{huang2021variational}
Chin-Wei Huang, Jae~Hyun Lim, and Aaron~C Courville.
\newblock A variational perspective on diffusion-based generative models and score matching.
\newblock \emph{NeurIPS}, 2021.

\bibitem[Jordan et~al.(1999)Jordan, Ghahramani, Jaakkola, and Saul]{jordan1999introduction}
Michael~I Jordan, Zoubin Ghahramani, Tommi~S Jaakkola, and Lawrence~K Saul.
\newblock An introduction to variational methods for graphical models.
\newblock \emph{Machine learning}, 37\penalty0 (2):\penalty0 183--233, 1999.

\bibitem[Kingma \& Welling(2014)Kingma and Welling]{kingma2013auto}
Diederik~P. Kingma and Max Welling.
\newblock Auto-encoding variational bayes.
\newblock In \emph{ICLR}, 2014.

\bibitem[Kingma et~al.(2021)Kingma, Salimans, Poole, and Ho]{kingma2021variational}
Diederik~P Kingma, Tim Salimans, Ben Poole, and Jonathan Ho.
\newblock Variational diffusion models.
\newblock In \emph{NeurIPS}, 2021.

\bibitem[Kingma et~al.(2016)Kingma, Salimans, Jozefowicz, Chen, Sutskever, and Welling]{kingma2016improved}
Durk~P Kingma, Tim Salimans, Rafal Jozefowicz, Xi~Chen, Ilya Sutskever, and Max Welling.
\newblock Improved variational inference with inverse autoregressive flow.
\newblock \emph{NeurIPS}, 2016.

\bibitem[Kuzina et~al.(2022)Kuzina, Welling, and Tomczak]{kuzina2022alleviating}
Anna Kuzina, Max Welling, and Jakub~Mikolaj Tomczak.
\newblock Alleviating adversarial attacks on variational autoencoders with mcmc.
\newblock In \emph{NeurIPS}, 2022.

\bibitem[Lucas et~al.(2019)Lucas, Tucker, Grosse, and Norouzi]{lucas2019understanding}
James Lucas, George Tucker, Roger Grosse, and Mohammad Norouzi.
\newblock Understanding posterior collapse in generative latent variable models.
\newblock \emph{Deep Generative Models for Highly Structured Data\@ICLR}, 2019.

\bibitem[Maal{\o}e et~al.(2017)Maal{\o}e, Fraccaro, and Winther]{maaloe2017semi}
Lars Maal{\o}e, Marco Fraccaro, and Ole Winther.
\newblock Semi-supervised generation with cluster-aware generative models.
\newblock \emph{arXiv}, 2017.

\bibitem[Maal{\o}e et~al.(2019)Maal{\o}e, Fraccaro, Li{\'e}vin, and Winther]{maaloe2019biva}
Lars Maal{\o}e, Marco Fraccaro, Valentin Li{\'e}vin, and Ole Winther.
\newblock Biva: A very deep hierarchy of latent variables for generative modeling.
\newblock \emph{NeurIPS}, 2019.

\bibitem[Nalisnick et~al.(2019)Nalisnick, Matsukawa, Teh, Gorur, and Lakshminarayanan]{nalisnick2018deep}
Eric Nalisnick, Akihiro Matsukawa, Yee~Whye Teh, Dilan Gorur, and Balaji Lakshminarayanan.
\newblock Do deep generative models know what they don't know?
\newblock In \emph{ICLR}, 2019.

\bibitem[Pennebaker \& Mitchell(1992)Pennebaker and Mitchell]{pennebaker1992jpeg}
William~B Pennebaker and Joan~L Mitchell.
\newblock \emph{{JPEG: Still image data compression standard}}.
\newblock Springer Science \& Business Media, 1992.

\bibitem[Perugachi-Diaz et~al.(2021)Perugachi-Diaz, Tomczak, and Bhulai]{perugachi2021invertible}
Yura Perugachi-Diaz, Jakub Tomczak, and Sandjai Bhulai.
\newblock Invertible densenets with concatenated lipswish.
\newblock \emph{NeurIPS}, 2021.

\bibitem[Pervez \& Gavves(2021)Pervez and Gavves]{pervez21a}
Adeel Pervez and Efstratios Gavves.
\newblock Spectral smoothing unveils phase transitions in hierarchical variational autoencoders.
\newblock \emph{ICML}, 2021.

\bibitem[Raue et~al.(2009)Raue, Kreutz, Maiwald, Bachmann, Schilling, Klingm{\"u}ller, and Timmer]{raue2009structural}
Andreas Raue, Clemens Kreutz, Thomas Maiwald, Julie Bachmann, Marcel Schilling, Ursula Klingm{\"u}ller, and Jens Timmer.
\newblock Structural and practical identifiability analysis of partially observed dynamical models by exploiting the profile likelihood.
\newblock \emph{Bioinformatics}, 25\penalty0 (15):\penalty0 1923--1929, 2009.

\bibitem[Razavi et~al.(2019)Razavi, van~den Oord, Poole, and Vinyals]{razavi2019preventing}
Ali Razavi, Aaron van~den Oord, Ben Poole, and Oriol Vinyals.
\newblock Preventing posterior collapse with delta-vaes.
\newblock In \emph{ICLR}, 2019.

\bibitem[Rezende et~al.(2014)Rezende, Mohamed, and Wierstra]{rezende2014stochastic}
Danilo~Jimenez Rezende, Shakir Mohamed, and Daan Wierstra.
\newblock Stochastic backpropagation and approximate inference in deep generative models.
\newblock In \emph{ICML}, 2014.

\bibitem[Sinha \& Dieng(2021)Sinha and Dieng]{sinha2021consistency}
Samarth Sinha and Adji~Bousso Dieng.
\newblock Consistency regularization for variational auto-encoders.
\newblock \emph{NeurIPS}, 2021.

\bibitem[Sohl-Dickstein et~al.(2015)Sohl-Dickstein, Weiss, Maheswaranathan, and Ganguli]{sohl2015deep}
Jascha Sohl-Dickstein, Eric Weiss, Niru Maheswaranathan, and Surya Ganguli.
\newblock Deep unsupervised learning using nonequilibrium thermodynamics.
\newblock In \emph{ICML}, 2015.

\bibitem[S{\o}nderby et~al.(2016)S{\o}nderby, Raiko, Maal{\o}e, S{\o}nderby, and Winther]{sonderby2016ladder}
Casper~Kaae S{\o}nderby, Tapani Raiko, Lars Maal{\o}e, S{\o}ren~Kaae S{\o}nderby, and Ole Winther.
\newblock Ladder variational autoencoders.
\newblock \emph{NeurIPS}, 2016.

\bibitem[Tomczak \& Welling(2018)Tomczak and Welling]{tomczak2018vae}
Jakub Tomczak and Max Welling.
\newblock Vae with a vampprior.
\newblock In \emph{AISTATS}, 2018.

\bibitem[Tomczak(2022)]{tomczak2022deep}
Jakub~M. Tomczak.
\newblock \emph{Deep Generative Modeling}.
\newblock Springer Cham, 2022.

\bibitem[Tzen \& Raginsky(2019)Tzen and Raginsky]{tzen2019neural}
Belinda Tzen and Maxim Raginsky.
\newblock Neural stochastic differential equations: Deep latent gaussian models in the diffusion limit.
\newblock \emph{arXiv}, 2019.

\bibitem[Vahdat \& Kautz(2020)Vahdat and Kautz]{vahdat2020nvae}
Arash Vahdat and Jan Kautz.
\newblock Nvae: A deep hierarchical variational autoencoder.
\newblock \emph{NeurIPS}, 2020.

\bibitem[Vahdat et~al.(2021)Vahdat, Kreis, and Kautz]{vahdat2021score}
Arash Vahdat, Karsten Kreis, and Jan Kautz.
\newblock Score-based generative modeling in latent space.
\newblock \emph{NeurIPS}, 2021.

\bibitem[Van Den~Oord et~al.(2017)Van Den~Oord, Vinyals, et~al.]{van2017neural}
Aaron Van Den~Oord, Oriol Vinyals, et~al.
\newblock Neural discrete representation learning.
\newblock \emph{NeurIPS}, 2017.

\bibitem[Wang et~al.(2021)Wang, Blei, and Cunningham]{wang2021posterior}
Yixin Wang, David Blei, and John~P Cunningham.
\newblock Posterior collapse and latent variable non-identifiability.
\newblock \emph{NeurIPS}, 2021.

\bibitem[Wang et~al.(2003)Wang, Simoncelli, and Bovik]{wang2003multiscale}
Zhou Wang, Eero~P Simoncelli, and Alan~C Bovik.
\newblock Multiscale structural similarity for image quality assessment.
\newblock In \emph{IEEE Conf. on Signals, Systems \& Computers}, 2003.

\bibitem[Wehenkel \& Louppe(2021)Wehenkel and Louppe]{wehenkel2021diffusion}
Antoine Wehenkel and Gilles Louppe.
\newblock Diffusion priors in variational autoencoders.
\newblock In \emph{INNF\@ICML}, 2021.

\end{thebibliography}
\bibliographystyle{iclr2024_conference}
\newpage
\appendix

\section{Posterior collapse and variational distribution}\label{appdx:variational_collapse}
Here, we present a discussion on the variational posterior collapse. To keep the notation uncluttered, we use $\rvz$ instead of $\rvz_{1:L}$. First, let us look into the Kullback-Leibler divergence between the variational posterior and the real posterior:
\begin{align*}
  D_{\mathrm{KL}}[q_{\phi}(\rvz|\rvx) || p_{\theta}(\rvz|\rvx)]  =& \int \Enc{\rvz}{\rvx} \ln \frac{\Enc{\rvz}{\rvx}}{p_{\theta}(\rvz|\rvx)} \mathrm{d}\rvz \\ 
     =& \int \Enc{\rvz}{\rvx} \ln \Enc{\rvz}{\rvx} \mathrm{d}\rvz - \int \Enc{\rvz}{\rvx} \ln p_{\theta}(\rvz|\rvx) \mathrm{d}\rvz \\
     =& -\mathbb{H}[\Enc{\rvz}{\rvx}] - \E_{\Enc{\rvz}{\rvx}} [ \ln p_{\theta}(\rvz|\rvx) ] \\
     =& -\mathbb{H}[\Enc{\rvz}{\rvx}] - \E_{\Enc{\rvz}{\rvx}} \left[ \ln\frac{ p_{\theta}(\rvx|\rvz) p_{\theta}(\rvz)}{p_{\theta}(\rvx)} \right] \\
     =& -\mathbb{H}[\Enc{\rvz}{\rvx}] - \E_{\Enc{\rvz}{\rvx}} \left[ \ln p_{\theta}(\rvx|\rvz) \right] - \E_{\Enc{\rvz}{\rvx}} \left[ \ln p_{\theta}(\rvz) \right] + \E_{\Enc{\rvz}{\rvx}} \left[ \ln p_{\theta}(\rvx) \right] \\
     =& \ln p(\rvx) - \mathcal{L}(\phi, \theta ;\rvx) . 
\end{align*}
In other words, the the Kullback-Leibler divergence between the variational posterior and the real posterior calculated is equal to the difference between the true marginal likelihood and the ELBO. Now, if we assume the variational posterior collapses, i.e., $q(\rvz | \rvx) = p(\rvz)$, then we get:
\begin{align*}
  D_{\mathrm{KL}}[q_{\phi}(\rvz|\rvx) || p_{\theta}(\rvz|\rvx)]  =& D_{\mathrm{KL}}[p(\rvz) || p_{\theta}(\rvz|\rvx)]  \\
     =& -\mathbb{H}[p(\rvz)] - \E_{p(\rvz)} \left[ \ln p_{\theta}(\rvx|\rvz) \right] - \E_{p(\rvz)} \left[ \ln p_{\theta}(\rvz) \right] + \E_{p(\rvz)} \left[ \ln p_{\theta}(\rvx) \right] \\
    =& -\mathbb{H}[p(\rvz)] - \E_{p(\rvz)} \left[ \ln p_{\theta}(\rvx|\rvz) \right] + \mathbb{H}[p(\rvz)] +  \ln p_{\theta}(\rvx) \\
    =&  \ln p_{\theta}(\rvx) - \E_{p(\rvz)} \left[ \ln p_{\theta}(\rvx|\rvz) \right]
\end{align*}

As a result, the gap between the collapsed variational posterior ($q_{\phi}(\rvz | \rvx) = p_{\theta}(\rvz)$) and the true posterior is equal to the difference between the marginal likelihood and $\E_{p(\rvz)} \left[ \ln p_{\theta}(\rvx|\rvz) \right]$. 

We can consider two cases, that is:
\begin{enumerate}
    \item If the real posterior collapses, $p_{\theta}(\rvz | \rvx) = p_{\theta}(\rvz)$, then naturally the variational posterior collapses. The reason is straightforward: We optimize the following objective: $D_{\mathrm{KL}}[q_{\phi}(\rvz|\rvx) || p_{\theta}(\rvz)]$.
    \item If the variational posterior collapses, then depending on the expressive power of the conditional likelihood $p_{\theta}(\rvx|\rvz)$, the true posterior can also collapse. This follows from the fact that if $\E_{p(\rvz)} \left[ \ln p_{\theta}(\rvx|\rvz) \right] \approx \ln p(\rvx)$, then $D_{\mathrm{KL}}[p(\rvz) || p_{\theta}(\rvz|\rvx)] = 0$, thus, $p_{\theta}(\rvz|\rvx) \approx p(\rvz)$.
\end{enumerate}

In the second point, it is still possible that the variational posterior can collapse and still the real posterior is not collapsed (or it is "partially" collapsed, meaning that it gets closer to $p_{\theta}(\rvz)$). 

\newpage
\section{Posterior collapse and latent variables non-identifiability}\label{appx:theory_proof}
\begin{propagain}{prop:collapse_cond_indep}
Consider a top-down hierarchical VAE introduced in Section \ref{sec:hierarchical_vae}. Then, for a given set of parameter values $\theta^*$, the posterior of the latent variable $\rvz_l$ collapses if and only if $\rvx$ and $\rvz_l$ are conditionally independent given ($\rvz_{l+1}, \dots, \rvz_L$).
\end{propagain}

\textit{Proof.} 
To simplify the notation, let us split the latent variables of hierarchical VAEs into three groups:
\begin{align} \label{eq:three_groups_app}
    \underbrace{\rvz_1, \dots, \rvz_{l-1}}_{\rvz_A},  \underbrace{\rvz_l}_{\rvz_B},  \underbrace{\rvz_{l+1}, \dots, \rvz_L}_{\rvz_C}.
\end{align}
We can do this for each $l \in {1, \dots, L}$, assuming that in the corner case of $l=1$, $\rvz_A$ is an empty set, and in the case of $l=L$, $\rvz_C$ is an empty set. 
Then, the posterior collapse implies $p_{\theta^*}(\rvz_B|\rvz_C, \rvx) = p_{\theta^*}(\rvz_B|\rvz_C)$, and the conditional independence is exactly the following equality: $p_{\theta^*}(\rvx| \rvz_B, \rvz_C) = p_{\theta^*}(\rvx|\rvz_C)$.
The proof follows directly from Theorem 1 in \citep{wang2021posterior}, where everything is additionally conditioned on $\rvz_C$. 
$\hfill\blacksquare$

Note, however, that the conditional independence in Proposition \ref{prop:collapse_cond_indep} is not the same as the latent variable non-identifiability which is defined as follows:

\begin{align}\label{eq:hierarchical_non_identifibility}
    p_{\theta^*} (\rvx | \rvz_{1:L}) =  p_{\theta^*} (\rvx | \rvz_{-l}),
\end{align}
where $\rvz_{-l} = (\rvz_1, \dots, \rvz_{l-1},\rvz_{l+1}, \dots, \rvz_L)$. To see how latent variable non-identifiability is connected to posterior collapse (Eq.~\ref{eq:posterior_collapse_definition}) in hierarchical VAE, we start with the following proposition.

\begin{propagain}{prop:cond_indep_to_identifiability}
Consider a top-down hierarchical VAE introduced in Section \ref{sec:hierarchical_vae}. If  $\rvx$ and $\rvz_l$ are conditionally independent given ($\rvz_{l+1}, \dots, \rvz_L$), then the latent variable $\rvz_l$ is non-identifiable. However, if $\rvz_l$ is non-identifiable, it does not imply that it is conditionally independent with $\rvx$ given ($\rvz_{l+1}, \dots, \rvz_L$).
\end{propagain}
\textit{Proof.} 
Let us utilize the same notation as in Proposition \ref{prop:collapse_cond_indep}. Consider conditional independence, namely $p_{\theta^*}(\rvx| \rvz_B, \rvz_C) = p_{\theta^*}(\rvx|\rvz_C)$. In other words, if we consider a corresponding graphical model, all the paths from $\rvz_B$ to $\rvx$ should go through $\rvz_C$. Then, for any $\rvz_A$ it holds that $p_{\theta^*}(\rvx| \rvz_A, \rvz_B, \rvz_C) = p_{\theta^*}(\rvx|\rvz_A, \rvz_C)$. This can be proved by contradiction. If this is not true, then there exists a path from $\rvz_B$ to $\rvx$, which does not go through $\rvz_A$ or $\rvz_C$. Therefore, there exists a path from $\rvz_B$ to $\rvx$, which does not go through $\rvz_C$. This contradicts the initial assumption. In summary, we have shown that if $\rvz_l$ and $\rvx$ are conditionally independent given ($\rvz_{l+1}, \dots \rvz_L$), then they are also conditionally independent given ($\rvz_{l}, \dots \rvz_{l-1}, \rvz_{l+1}, \dots \rvz_L$), which is the definition of the latent variable non-identifiability. 

\begin{wrapfigure}{r}{0.5\textwidth}
\vskip -10pt
    \centering
    \begin{tikzpicture}[node distance=1.cm]
  \node[latent, minimum size=0.5cm] (C) {$\rvz_C$};
  \node[latent, minimum size=0.5cm, right=of C] (B) {$\rvz_B$};
  \node[latent, minimum size=0.5cm, right=of B] (A) {$\rvz_A$};
  \node[latent, minimum size=0.5cm, right=of A] (X) {$\rvx$};
  
  \edge[-, thick] {C} {B};
  \edge[-, thick] {B} {A};
  \edge[-, thick] {A} {X};
  \draw [-, thick, bend left] (C) to node[midway,above] {} (X);
\end{tikzpicture}
    \caption{Example of a graphical model where $\rvz_B$ and $\rvx$ are conditionally independent given $\rvz_A, \rvz_C$ (non-idetifiability). However, they are not conditionally independent given only $\rvz_C$, since there is an additional path from $\rvz_B$ to $\rvx$ through $\rvz_A$.}
    \label{fig:graph_model_example}
\vskip -7pt
\end{wrapfigure}
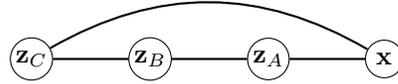
To see that the opposite is false, consider the counter example in Figure \ref{fig:graph_model_example}. In this graphical model, $\rvz_B$ and $\rvx$ are conditionally independent given $\rvz_A, \rvz_C$. Namely, all the paths from $\rvz_B$ to $\rvx$ go through either $\rvz_A$ or $\rvz_C$. However, if we are given only $\rvz_C$, there is still a path from $\rvz_B$ to $\rvx$ (going through $\rvz_A$). Therefore, $\rvz_B$ and $\rvx$ are not conditionally independent given $\rvz_C$. This implies that the latent variable non-identifiability does not imply conditional independence. 
$\hfill\blacksquare$


\newpage
\section{Background of diffusion probabilistic models}\label{appendix:ddgm_theory}

\textit{Diffusion Probabilistic Models}  or Diffusion-based Deep Generative Models \citep{ho2020denoising, sohl2015deep} constitute a class of generative models that can be viewed as a special case of the Hierarchical VAEs \citep{huang2021variational, kingma2021variational, tomczak2022deep, tzen2019neural}. 

Denoting the last latent (context) $\rvz_{L} \equiv \rvy_0$ and auxiliary latent variables $\rvy_{t}$, $t = 1, \ldots, T$, we define a generative model, also referred to as the \textit{backward} (or \textit{reverse}) \textit{process}, as a Markov chain with Gaussian transitions starting with $p(\rvy_T) = \mathcal{N}(\rvy_{T}| \boldsymbol{0}, \mathbf{I})$, that is: $p_{\gamma}(\rvy_0, \ldots, \rvy_{T}) = p(\rvy_T)\ \prod_{t=0}^{T} p_{\gamma}(\rvy_{t-1} | \rvy_{t})$, where $p_{\gamma}(\rvy_{t-1} | \rvy_{t}) = \mathcal{N}(\rvy_{t-1}; \mu_{\gamma}(\rvy_{t}, t), \Sigma_{\gamma}(\rvy_{t}, t))$.

Let us further define $\alpha_t = 1 - \beta_t$ and $\overline{\alpha}_t = \prod_{i=0}^{t} \alpha_{i}$. Since the conditionals in the forward diffusion can be seen as Gaussian linear models, we can analytically calculate the following distributions: 
\begin{align}\label{eq:forward_diffusion_latent_from_data}
    q(\rvy_t|\rvy_0) = &\mathcal{N}(\rvy_t; \sqrt{\overline{\alpha}_t} \rvy_0, (1 - \overline{\alpha}_t) \mathbf{I}), \\
    \label{eq:forward_diffusion_previous_latent}
q(\rvy_{t-1}|\rvy_{t}, \rvx_0) = &\mathcal{N}(\rvy_{t-1}; \Tilde{\mu}(\rvy_t, \rvy_0), \Tilde{\beta}_t \mathbf{I}) ,
\end{align}
where $\Tilde{\mu}(\rvy_t, \rvy_0) = \frac{\sqrt{\overline{\alpha}_{t-1}} \beta_{t}}{1-\overline{\alpha}_{t}} \rvy_0 + \frac{\sqrt{\alpha_{t}}\left(1-\overline{\alpha}_{t-1}\right)}{1-\overline{\alpha}_{t}} \rvy_{t}$, 
and $\Tilde{\beta}_{t}=\frac{1-\overline{\alpha}_{t-1}}{1-\overline{\alpha}_{t}} \beta_{t}$. We can use (\ref{eq:forward_diffusion_latent_from_data}) and (\ref{eq:forward_diffusion_previous_latent}) to define the variational lower bound as follows:
\begin{align} \label{eq:diff_elbo}
    L_{vlb} = & \underbrace{\E_{q(\rvy_1|\rvy_0)}[\ln p_{\gamma}(\rvy_0|\rvy_1)]}_{-L_0} 
    - \underbrace{\KL{q(\rvy_T |\rvy_0)}{p(\rvy_T)}}_{L_T}  \\
    & - \sum_{t=2}^T \underbrace{\E_{q(\rvy_t|\rvy_0)} \KL{q(\rvy_{t-1}|\rvy_t, \rvy_0)}{p_{\gamma}(\rvy_{t-1}|\rvy_t)}}_{L_{t-1}}. \notag
\end{align}
Parameters $\gamma$ of the diffusion model and parameters $\theta, \phi$ of the hierarchical VAE are optimized simultaneously with the joint objective Eq. \ref{eq:our_elbo}, where we use the lower bound (Eq. \ref{eq:diff_elbo}) instead of the $\ln p_{\gamma}(f(\rvx))$ term.

\textit{The conditional distribution over the context} We assume that the context is a discrete random variable. Therefore, it is important to choose an appropriate family of conditional distributions $p_{\gamma}(\rvy_0 | \rvy_1)$. Following \cite{ho2020denoising}, we scale $\rvy_0$ linearly to $[-1, 1]$, and use the discretized (binned) Gaussian distribution:
\begin{equation}\label{eq:discretized_Gaussian}
    p_{\gamma}\left(\rvy_{0} | \rvy_{1}\right)=\prod_{i=1}^{D} \int_{\delta_{-}\left(x_{0}^{i}\right)}^{\delta_{+}\left(x_{0}^{i}\right)} \mathcal{N}\left(x ; \mu_{\gamma}^{i}\left(\mathbf{x}_{1}, 1\right), \sigma_{1}^{2}\right) \mathrm{d} x, 
\end{equation}
where $D$ is the dimensionality of $\rvy_0$, and $i$ denotes one coordinate of $\rvy_0$, and:
\begin{equation}
    \delta_{+}(x)=\left\{\begin{array}{ll}
\infty & \text { if } x=1 \\
x+\frac{1}{b} & \text { if } x<1
\end{array} \quad \delta_{-}(x)= \begin{cases}-\infty & \text { if } x=-1 \\
x-\frac{1}{b} & \text { if } x>-1\end{cases}\right. ,
\end{equation}
where $b$ is the bin width determined based on training data.
\newpage
\section{CIFAR10 experiments}\label{appx:cifar_full}

In addition to the binary datasets, we perform experiments on natural images. We used the CIFAR10 dataset, which is a common benchmark in VAE literature. We report the results in n Table \ref{tab:main_results}. 
We observe that our approach works on par with the generative models which have comparable sizes (OU-VAE, Residual Flows, GLOW). However, there are models with much larger sizes (e.g. VDVAE, NVAE), which perform better. Unfortunately, we do not have the computational resources to train a comparable-size model. Instead, we compare the DCT-VAE with our implementation of the smaller-size VDVAE. 

\begin{table}[ht]
    \centering
    \caption{The test performance on CIFAR10 dataset. We compare the total number of trainable parameters (Size), the number of stochastic layers ($L$), and NLL.\\ \textsuperscript{$\dagger$} Results with data augmentation.}
    \vskip 5pt
    \label{tab:main_results}
    \begin{tabular}{l||rlc}
        \toprule
        \textsc{Model}  & 
         \textsc{Size} & \textsc{L} & \small{\textsc{bits/dim $ \leq \, \downarrow $}} \\
        \midrule
        \textbf{DCT-VAE (ours)}& 22M & 29 & 3.26  \\
        Small VDVAE    & \multirow{2}{*}{21M} & \multirow{2}{*}{29} & \multirow{2}{*}{3.28}  \\
        (our implementation) \\
        Attentive VAE & \multirow{2}{*}{119M} & \multirow{2}{*}{16} &   \multirow{2}{*}{2.79} \\
         \small{\citep{apostolopoulou2021deep}}\\
        VDVAE \small{\citep{child2020very}} & 39M & 45& 2.87 \\
        Residual flows& \multirow{2}{*}{25M} & \multirow{2}{*}{1}
                & \multirow{2}{*}{3.28}  \\
                 \small{\citep{perugachi2021invertible}}\\
        i-DenseNet flows & \multirow{2}{*}{25M} & \multirow{2}{*}{1}
                & \multirow{2}{*}{3.25}  \\
                \small{\citep{perugachi2021invertible}}\\
        OU-VAE  &  \multirow{2}{*}{10M} & \multirow{2}{*}{3} & \multirow{2}{*}{3.39}  \\
        \small{\citep{pervez21a}}\\
        CR-NVAE   & \multirow{2}{*}{131M} & \multirow{2}{*}{30}& \multirow{2}{*}{2.51\textsuperscript{$\dagger$}}   \\
        \small{\citep{sinha2021consistency}}\\
        NVAE &  ---   &     30  & 2.91     \\
         \small{\citep{vahdat2020nvae}} \\
        BIVA \small{\citep{maaloe2019biva}}  & 103M & 15 & 3.08  \\
        GLOW  & --- & 1
                & 3.46  \\
                \small{\citep{nalisnick2018deep}}\\
        IAF-VAE &--- & 12
                & 3.11  \\
                 \small{\citep{kingma2016improved}}\\
        \bottomrule
    \end{tabular}
\end{table} 

\newpage
\section{Model details}
\subsection{Architecture} \label{appendix:schema}
\begin{figure}
    \centering
        \begin{tabular}{c}
        \includegraphics[width=0.45\linewidth]{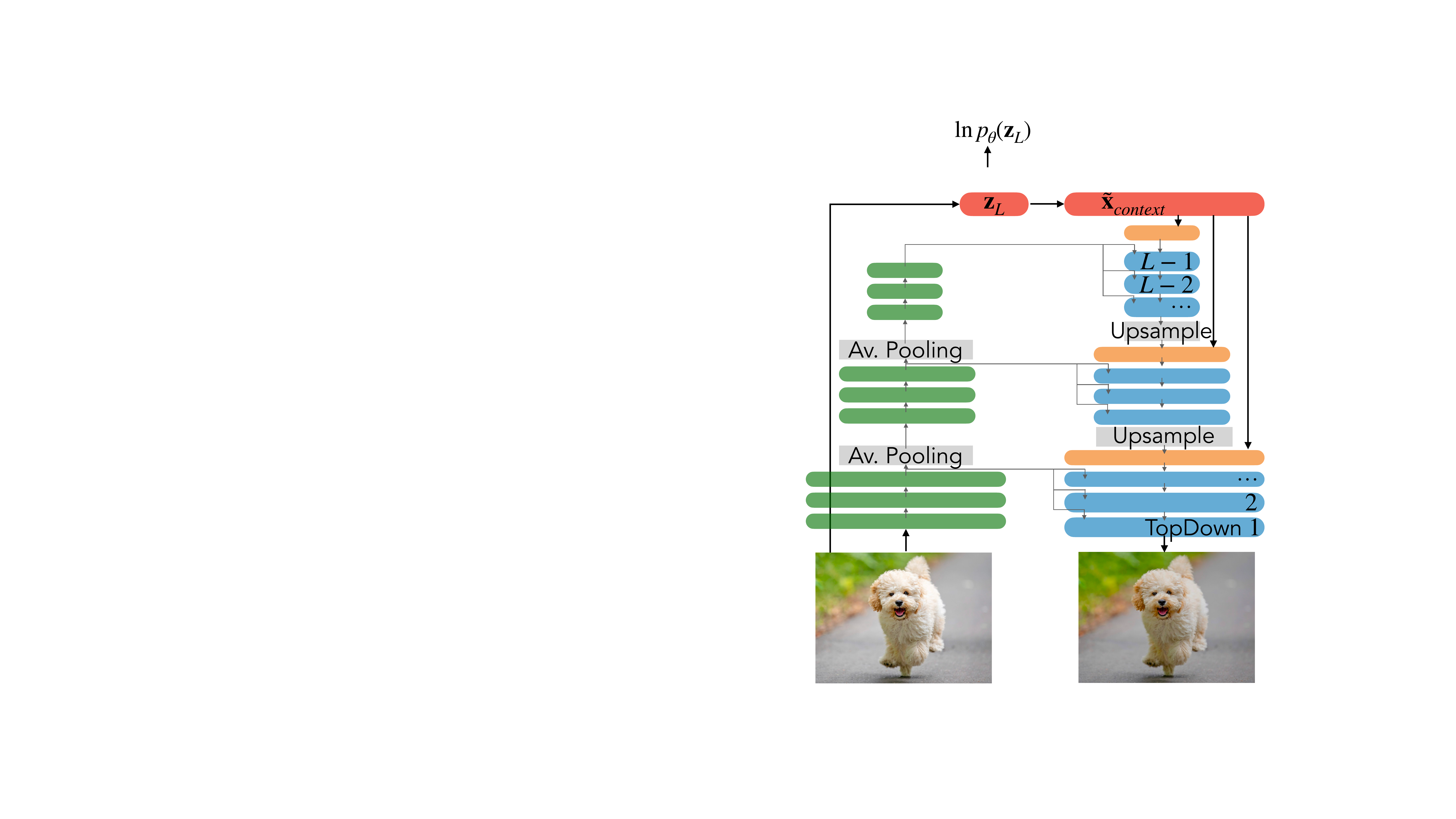} \\
        \includegraphics[width=0.50\linewidth]{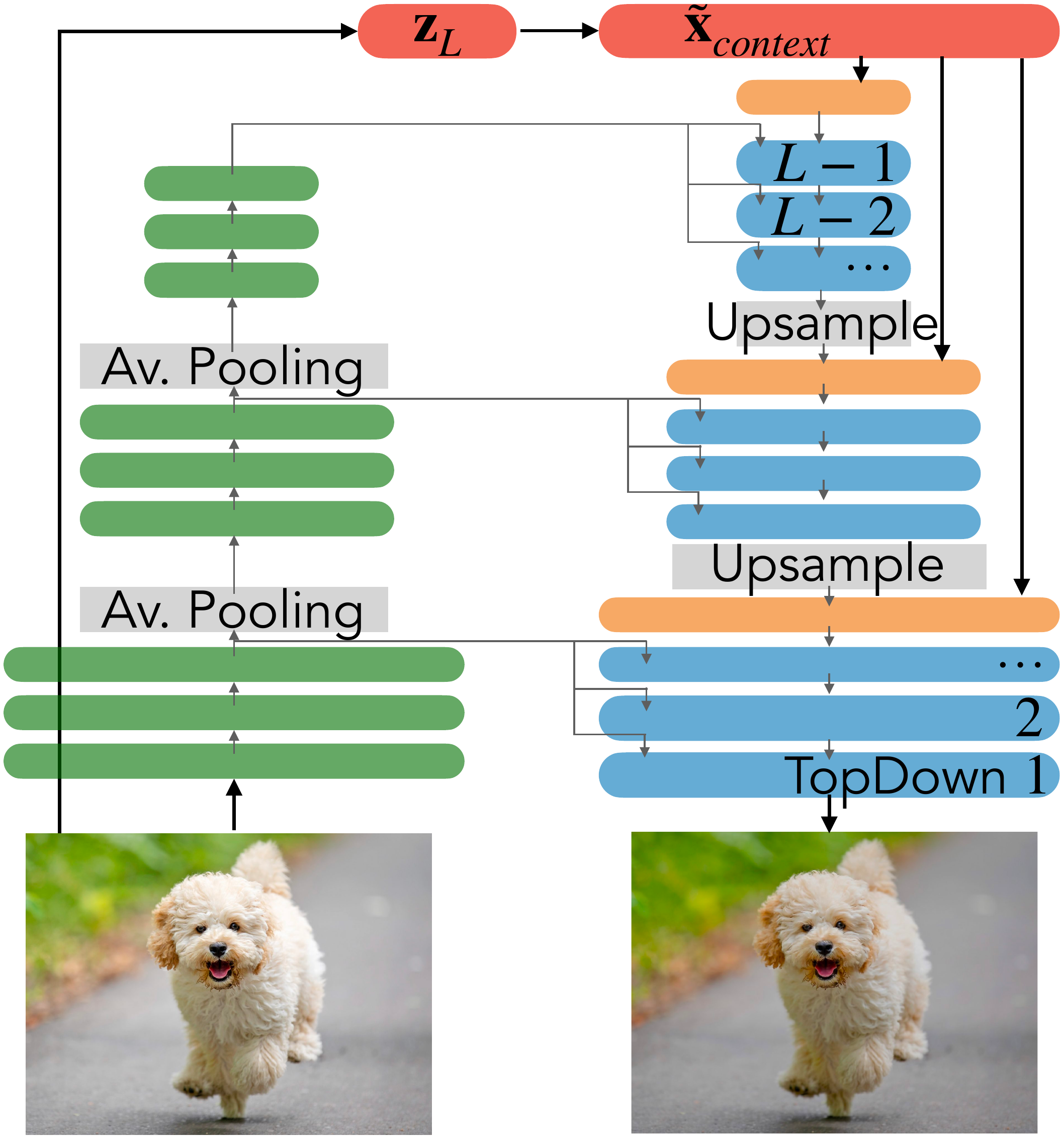} \\  
    \end{tabular}
    \caption{A diagram of the top-down hierarchical VAE with the context. The decoder consists of \textit{TopDown} blocks (blue), which take as input features from the block above $\mathbf{h}^{Dec}_{l+1}$ and the features from the encoder $\mathbf{h}^{Enc}_{l+1}$ (only during training). Dotted lines denote that $\rvz_l$ is a sample from the prior (in the generative mode) or from the variational posterior (in the reconstruction mode). The \textit{context} (in red) is added to the features of the decoder at the beginning of each scale. 
    The encoder consists of ResNet blocks (green)
    We use the same ResNet blocks in the TopDown blocks.
    }
    \label{fig:model_schema}
\end{figure}

We schematically depict the proposed deep hierarchical VAE in Figure \ref{fig:model_schema}. We extend the architecture presented in \citep{child2020very} by using a deterministic, non-trainable function to create latent variable $\rvz_L$ (the context). It is then used to train the prior $p_{\theta}(\rvz_L)$, and to obtain $\tilde{\rvx}_{context}$ that is eventually passed to every level (scale) of the top-down decoder.

\newpage
\subsection{Hyperparameters}\label{appendix:hyperparams}
In Table \ref{tab:setup}, we report all the hyperparameter values that were used to train the baseline VAE and DCT-VAE.

\paragraph{The context Prior}
We use the diffusion generative model as a prior over the context. As a backbone, we use UNet implementation from \citep{dhariwal2021diffusion} which is available on GitHub\footnote{https://github.com/openai/guided-diffusion} with the hyperparameters provided in Table \ref{tab:setup}.

\begin{table}
\caption{Full list of hyperparameters.}
\label{tab:setup}
\begin{center}
\resizebox{.99\textwidth}{!}{
\begin{tabular}{ll||cc|cc|cc}
\toprule
& &  \multicolumn{2}{c|}{MNIST}   & \multicolumn{2}{c|}{OMNIGLOT} 
& \multicolumn{2}{c}{CIFAR10} \\
& & VAE & DCT-VAE & VAE & DCT-VAE 
& VAE & DCT-VAE \\ \midrule
\small{\multirow{11}{*}{\STAB{\rotatebox[origin=c]{90}{Optimization}}}} 
& Optimizer              &  \multicolumn{2}{c|}{AdamW}  & \multicolumn{2}{c|}{AdamW}  
                         & \multicolumn{2}{c}{AdamW} 
\\
& Scheduler              &  \multicolumn{2}{c|}{Cosine} & \multicolumn{2}{c|}{Cosine} 
                         & \multicolumn{2}{c}{Cosine}
\\
& Starting Learning rate &  \multicolumn{2}{c|}{1e-3}   & \multicolumn{2}{c|}{1e-3} 
                         & \multicolumn{2}{c}{4e-4} 
\\ 
& End Learning rate      &  \multicolumn{2}{c|}{1e-5}   & \multicolumn{2}{c|}{1e-5} 
                         & \multicolumn{2}{c}{5e-5}
\\ 
& Weight Decay           &  \multicolumn{2}{c|}{1e-2}   & \multicolumn{2}{c|}{1e-2} 
                            & \multicolumn{2}{c}{1e-2}
\\
& \# Epochs              & \multicolumn{2}{c|}{600}     & \multicolumn{2}{c|}{600}
                         & \multicolumn{2}{c}{8000}
\\
& Grad. Clipping         & \multicolumn{2}{c|}{1}       & \multicolumn{2}{c|}{1}
                         & \multicolumn{2}{c}{0.2}
\\
& Grad. Skipping Threshold & \multicolumn{2}{c|}{100}   & \multicolumn{2}{c|}{100}
                           & \multicolumn{2}{c}{100}
\\
& EMA rate               & \multicolumn{2}{c|}{0}       & \multicolumn{2}{c|}{0}
                         & \multicolumn{2}{c}{0}
\\
& \# GPUs                & \multicolumn{2}{c|}{1}       & \multicolumn{2}{c|}{1}
                         & \multicolumn{2}{c}{4}
\\
& Batch Size (per GPU)   & \multicolumn{2}{c|}{128}     & \multicolumn{2}{c|}{128}
                         & \multicolumn{2}{c}{96}
\\ \midrule
\multirow{8}{*}{\STAB{\rotatebox[origin=c]{90}{Architecture}}} 
& L                             & \multicolumn{2}{c|}{8}              & \multicolumn{2}{c|}{8}
                                &  \multicolumn{2}{c}{29}
\\
& \multirow{3}{*}{Latent Sizes} & $4\times14^2$,     & $4\times14^2$, & $4\times14^2$, & $4\times14^2$,  
                                & $10\times 32^2$, & $10\times 32^2$,
\\
&                               & $4\times7^2$.   & $3\times7^2$.     & $4\times7^2$. & $3\times7^2$.  
                                & $10\times 16^2$, $5\times 8^2$,    & $10\times 16^2$, $5\times 8^2$, 
\\
&                               &                 &             &  &
                                &$3\times 4^2$, $1\times 1^2$.     & $2\times 4^2$, $1\times 1^2$.
\\
& Latent Width                  & \multicolumn{2}{c|}{1}              &  \multicolumn{2}{c|}{1} 
                                &  \multicolumn{2}{c}{8} 
\\
& Context Size                  & ---    & $1\times6\times6$              & ---    & $1\times7\times7$ 
                                & ---    & $3\times6\times6$ 
\\
& \# Channels (input)           &  \multicolumn{2}{c|}{32}            &  \multicolumn{2}{c|}{32} 
                                &  \multicolumn{2}{c}{384} 
\\
& \# Channels (hidden)          &  \multicolumn{2}{c|}{40}            & \multicolumn{2}{c|}{40}
                                &  \multicolumn{2}{c}{96} 
\\
& Weight Norm                   &  \multicolumn{2}{c|}{\textsc{False}}& \multicolumn{2}{c|}{\textsc{False}}
                                & \multicolumn{2}{c}{\textsc{True}}
\\
& Activation                    & \multicolumn{2}{c|}{SiLU}  & \multicolumn{2}{c|}{SiLU}
                                & \multicolumn{2}{c}{SiLU}
\\
& Likelihood                    & \multicolumn{2}{c|}{Bernoulli}  & \multicolumn{2}{c|}{Bernoulli}
                                & \multicolumn{2}{c}{Discretized Logisitc Mixture}
\\
\midrule
\small{\multirow{4}{*}{\STAB{\rotatebox[origin=c]{90}{Context Prior}}}} 
& \# Diffusion Steps     &  --- & 7      &  --- & 7      
                         & --- &  40\\
& \# Scales in UNet      &  --- & 1      &  --- & 1      
                         & --- &  2\\
& \# ResBlocks per Scale &  --- & 3      &  --- & 3      
                         & --- &  3\\
& \# Channels            &  --- & 32     &  --- & 32     
                         & --- &  64\\
& $\beta$ schedule       &  --- & linear &  --- & linear 
                         & --- &  linear\\
\bottomrule
\end{tabular}}
\end{center}
\end{table}

\newpage
\section{Downsampling-based Context}\label{appx:context_down}
In this work, we propose a DCT-based context. However, downsampling can also be used to create a lower-dimensional representation of the input. Therefore, we conducted an ablation study where we used downsampled-based context. Results of this experiment can be found in Section \ref{sect:exp_image_generations}. 

To create a downsampling-based context we use average pooling, as shown in Algorithm \ref{alg:context_down_forward}. Then, we can decode it back by simply using nearest-neighbours upsampling (Algorithm \ref{alg:context_down_backward}). 

\begin{minipage}[t]{0.46\textwidth}
\begin{algorithm}[H]
\caption{Create context: downsampling}
\label{alg:context_down_forward}
\begin{algorithmic}
        \State \hskip-3mm \textbf{Input}: $\rvx, v$
    \State $\rvz_{\text{Downsample}} = \text{Av. Pooling}(\rvx, v)$
        \State $\rvz_{\text{Downsample}} = \text{quantize}(\rvz_{\text{Downsample}})$
        \State  \hskip-3mm \textbf{Return}: $\rvz_{\text{Downsample}}$
\end{algorithmic}
\end{algorithm}
\end{minipage}
\hfill
\begin{minipage}[t]{0.47\textwidth}
\begin{algorithm}[H]
\caption{Decode context: downsampling}
\label{alg:context_down_backward}
\begin{algorithmic} 
        \State \hskip-3mm \textbf{Input}: $\rvz_{\text{Downsample}}, D$\\
        \Comment{Apply nearest neighbour upsampling}
        \State $\tilde{\rvx}_{context} = \text{Upsampling}(\rvz_{\text{Downsample}}, D)$ 
        \State \hskip-3mm \textbf{Return}: $\tilde{\rvx}_{context}$
\end{algorithmic}
\end{algorithm}
\end{minipage}

\newpage
\section{Compression} \label{appx:compression}
To find out how much information about the data is preserved in the top latent variable, we conduct an experiment
where we use the baseline VDVAE and the DCT-VAE pretrained on CIFAR10 for compression. We use the KODAK dataset, which is a standard compression benchmark containing 24 images with resolution $512 \times 768$. Since CIFAR10 images are $32 \times 32$, we independently encode patches of
KODAK images. We then reconstruct each patch using only a part of the latent variables and combine these patches to obtain final reconstructions.

\begin{figure*}[!htbp]
    \begin{tabular}{>{\centering}p{0.3\textwidth}>{\centering}p{0.3\textwidth}>{\centering\arraybackslash}p{0.3\textwidth}}
        \multicolumn{3}{c}{\includegraphics[width=0.85\textwidth]{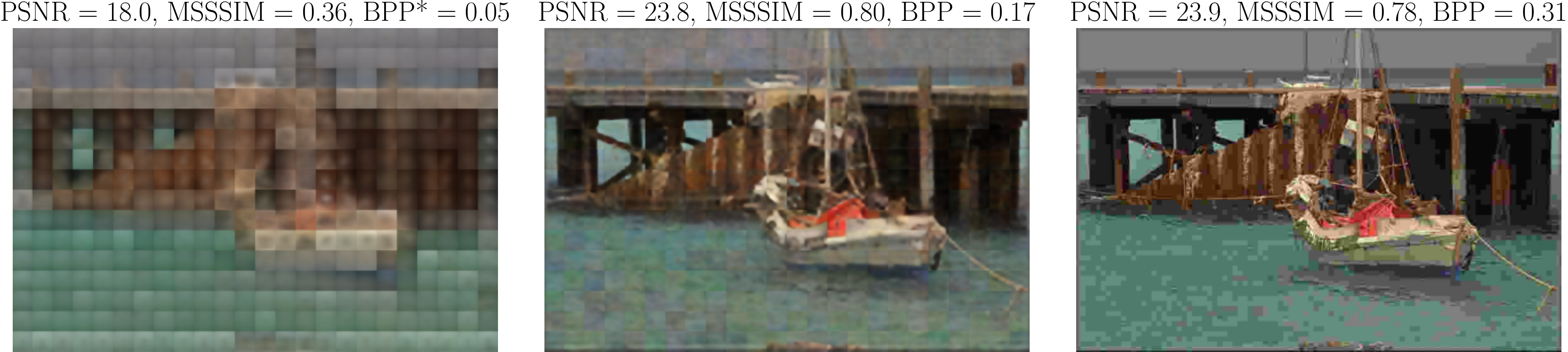} }
        \\
        \multicolumn{3}{c}{\includegraphics[width=0.85\textwidth]{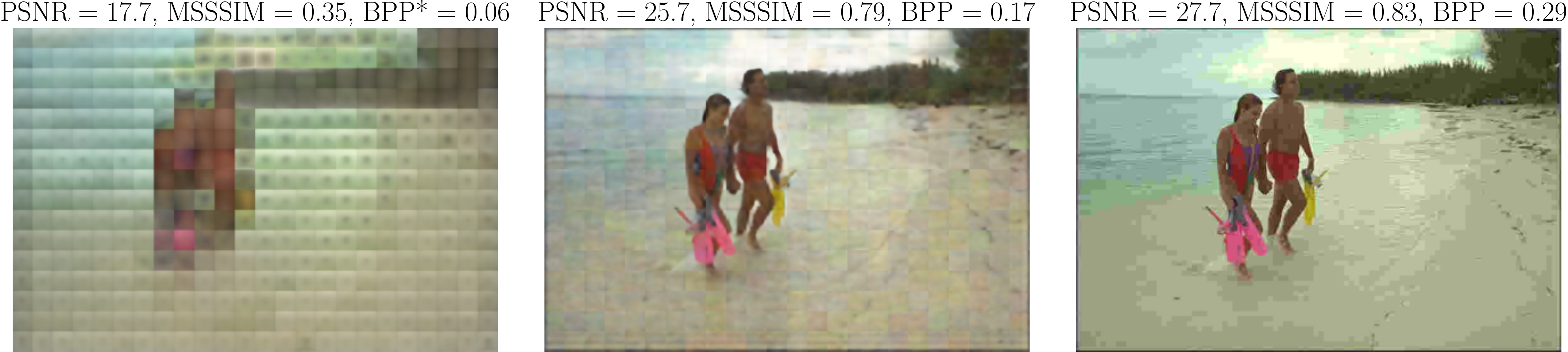} }
        \\
        \multicolumn{3}{c}{\includegraphics[width=0.85\textwidth]{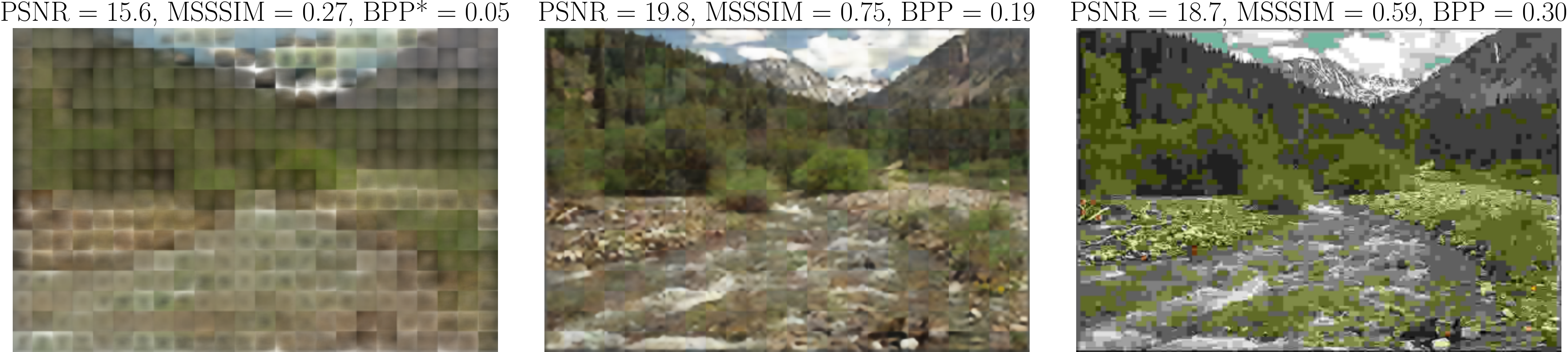} }
        \\
        \multicolumn{3}{c}{\includegraphics[width=0.85\textwidth]{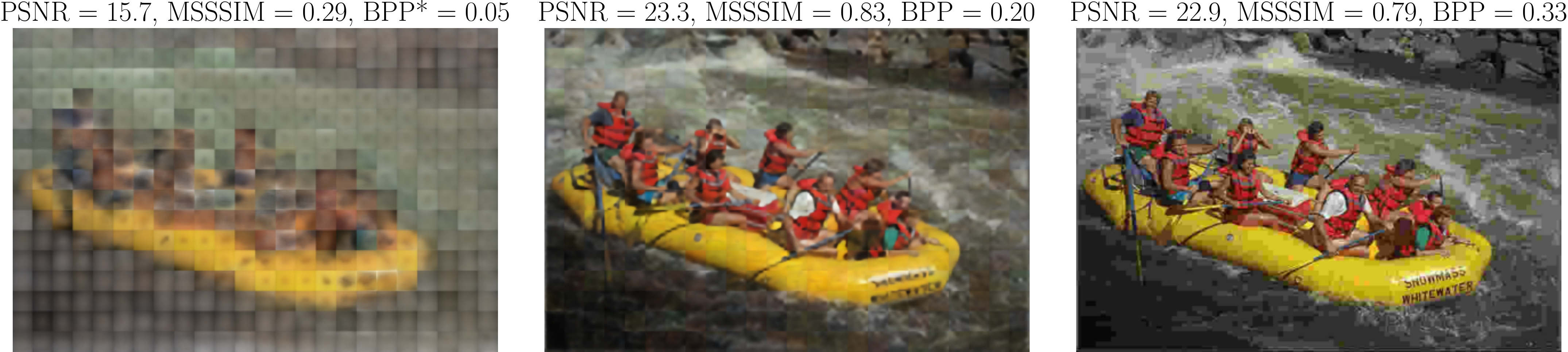} }
        \\
        \multicolumn{3}{c}{\includegraphics[width=0.85\textwidth]{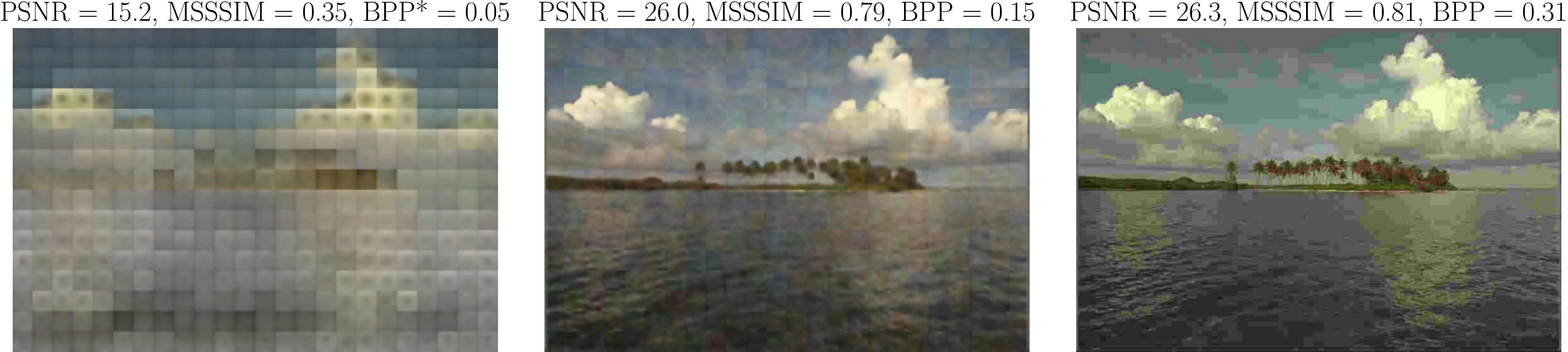} }
        \\
        \multicolumn{3}{c}{\includegraphics[width=0.85\textwidth]{pics/cifar10_kodak_example_14.pdf} } \\
        (a) VAE & (b) DCT-VAE  &(c) JPEG \\
        
    \end{tabular}
    \caption{Examples of the compressed images. We use 2 top latent variables of VDVAE to reconstruct the image in col. (a) and only context of DCT-VAE in col. (b). We choose JPEG compression to have a similar PSNR value to DCT-VAE (col. c).}
    \label{fig:cifar_kodak_example_extended}
\end{figure*}

In Figure \ref{fig:cifar_kodak_example_extended}, we present non-cherry-picked reconstructions from the compression experiment. We use a single latent variable (only context) for DCT-VAE and two top latent variables for the baseline model. We sample the rest of the latent variables from the prior distribution with a temperature equal to 0.1. We also show images compressed with JPEG for comparison. We use PSNR and MSSSIM to measure the reconstruction error. We report KL-divergence converted to bits-per-pixel as a compression rate. All latent variables (except for the context in DCT-VAE) are continuous.

\newpage

\newpage
\section{Robustness to Adversarial Attacks} \label{appx:adv_attack}
\begin{figure}[htbp]
    \begin{adjustbox}{center}
    \begin{tabular}{ccc}
    \includegraphics[width=0.3\textwidth]{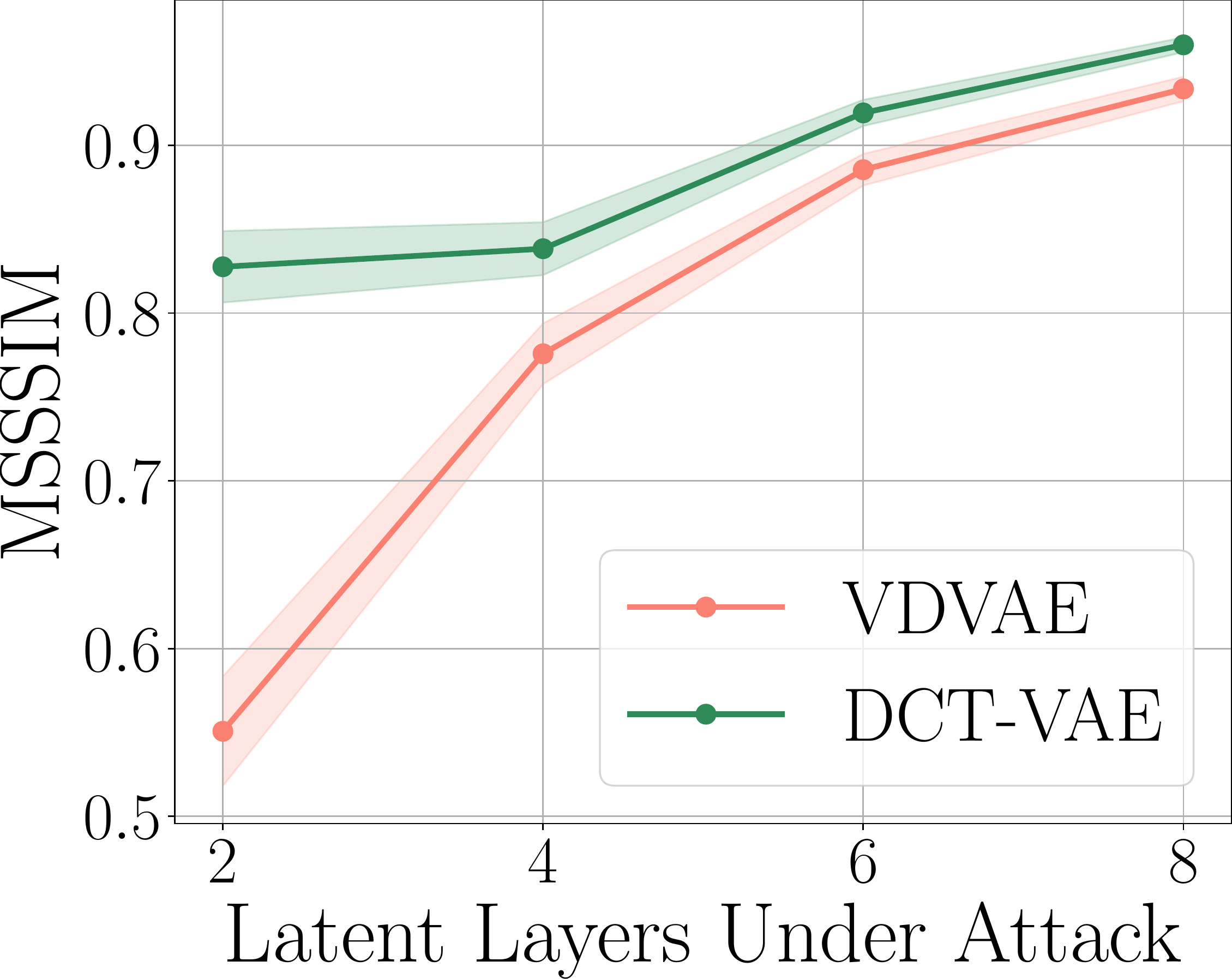} &
         \includegraphics[width=0.3\textwidth]{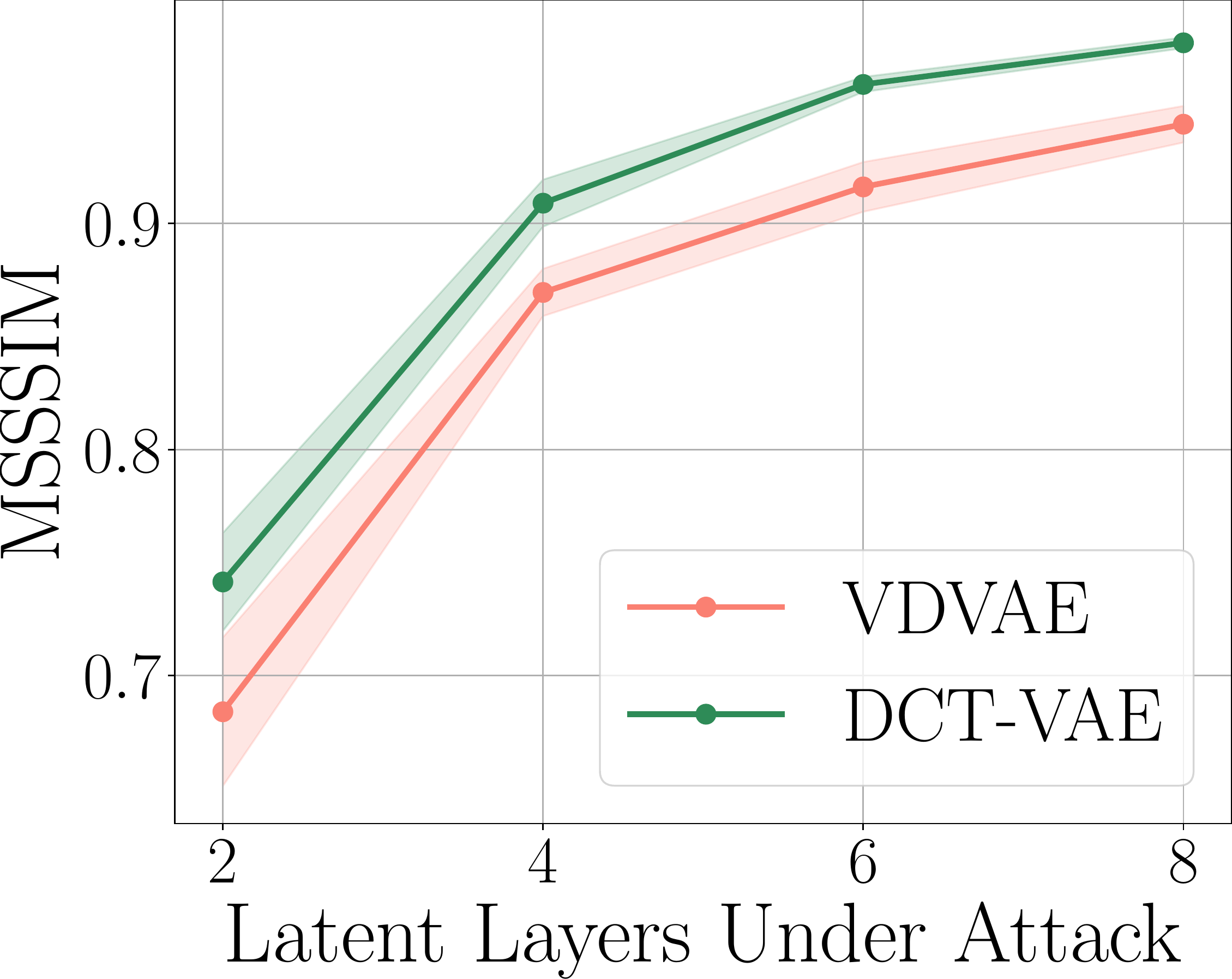} &
              \includegraphics[width=0.3\textwidth]{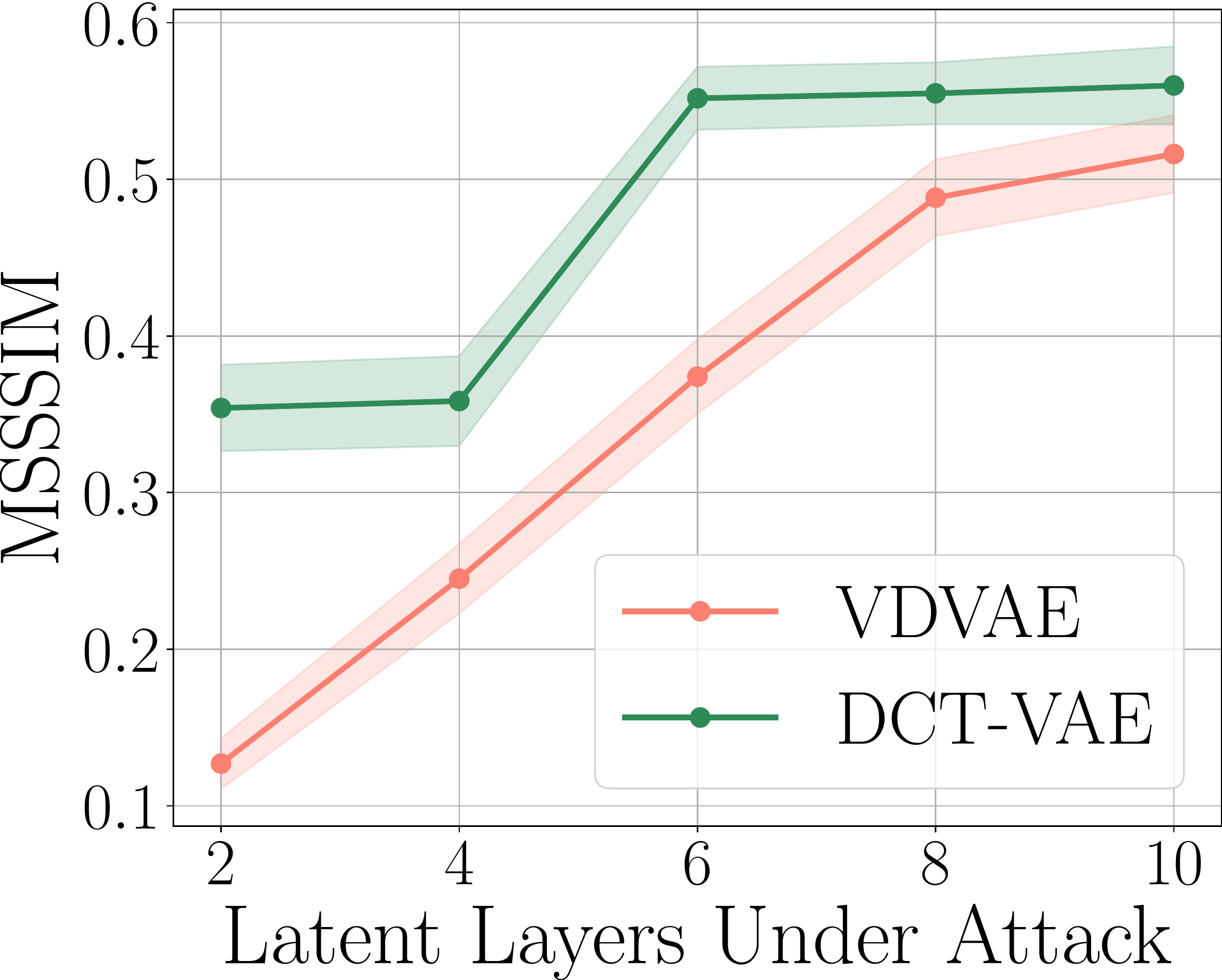} \\
         (a) MNIST & (b) OMNIGLOT& (c) CIFAR10\\
    \end{tabular}
    \end{adjustbox}
    \caption{The adversarial robustness measured by MSSSIM.}
    \label{fig:adversarial_results}
\end{figure}

In \citep{kuzina2022alleviating} it was shown that the top latent of deep hierarchical VAEs can be easily "fooled" by the most straightforward methods of attack construction, and thus, it could serve as a diagnostic tool to assess the robustness of the latent space. Here, we follow this line of thought to assess the robustness of the DCT-VAE. For each dataset, we use 50 test points (5 different random initializations) to construct latent space attacks on the VDVAE and the DCT-VAE. In Figure \ref{fig:adversarial_results}, we present the average similarity between the real reconstruction and the attacked reconstruction measured by MSSSIM depending on the latent layers under attack. In all cases, we see a clear advantage in using the DCT-based context. For MNIST and CIFAR10, the DCT-VAE provides much better robustness for the two latent layers under attack. In general, the DCT-VAE seems to be less affected by adversarial attacks than the VDVAE.

\end{document}